\documentclass[runningheads]{llncs}

\usepackage{eccv}

\usepackage{eccvabbrv}

\usepackage{graphicx}
\usepackage{booktabs}

\usepackage{tcolorbox}
\usepackage{amsmath}
\usepackage{subcaption}
\usepackage{multirow}
\usepackage{algorithm}
\usepackage{algpseudocode}
\usepackage{tikz}
\usepackage{bm}
\usepackage{xcolor}
\usepackage{adjustbox} 
\usepackage{float}
\usepackage{tabularx}
\usepackage[table]{xcolor}
\usepackage{nicematrix}
\usepackage[retain-zero-uncertainty=true]{siunitx}

\usetikzlibrary{arrows.meta, positioning, shapes.multipart, shapes.geometric, positioning, fit, calc, ext.paths.ortho, decorations.pathreplacing}
\usetikzlibrary{arrows}

\usepackage[accsupp]{axessibility}  

\usepackage{hyperref}

\usepackage{orcidlink}

\begin{document}

\title{\textsc{DeLux}: Cross-Modal Local Artifact Restoration in Video Using Neuromorphic Data} 

\titlerunning{DeLux: Cross-Modal Local Artifact Restoration}


\author{Bartosz Stachowiak\inst{1}\orcidlink{0009-0005-3518-9311} \and
Dariusz Brzezinski\inst{1}\orcidlink{0000-0001-9723-525X}}

\authorrunning{B. Stachowiak and D. Brzezinski}

\institute{Institute of Computing Science, Poznan University of Technology,\\
ul. Piotrowo 2, 60-965 Poznan, Poland\\
\email{\{bartosz.stachowiak,dariusz.brzezinski\}@cs.put.poznan.pl}}



\maketitle

\begin{figure}
    \centering
    \input{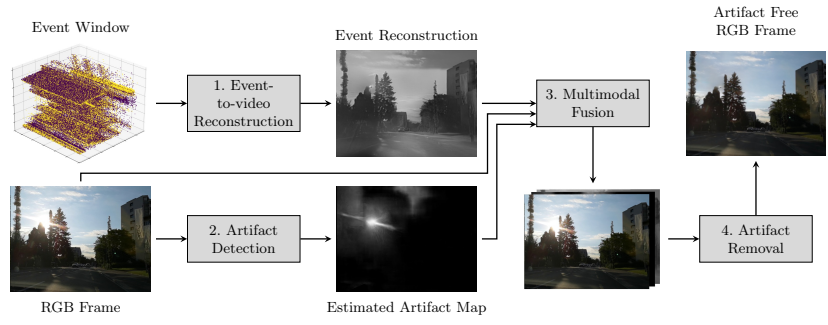}
    \caption{Overview of the proposed \textsc{DeLux} framework for cross-modal local artifact restoration. To address the challenge of localized image information loss, our modular pipeline explicitly decouples artifact detection from multimodal fusion and image inpainting. As illustrated, an RGB frame corrupted by glare is processed alongside a corresponding neuromorphic event window. By using event-to-video reconstruction as a prior and explicitly isolating the degraded region via an estimated artifact map, the multimodal fusion stage guides the targeted inpainting of corrupted RGB pixels.}
    \label{fig:our-pipeline}
\end{figure}

\begin{abstract}
Conventional RGB cameras suffer from lighting artifacts such as flare, glare, flicker, and overexposure, leading to irrecoverable information loss that necessitates computational restoration. However, existing approaches treat these problems in isolation, failing to recover structural details completely obscured by complex spatially discrete image degradations. In this paper, we propose a novel cross-modal restoration paradigm and present \textsc{DeLux}, a modular proof-of-concept pipeline that leverages neuromorphic event streams as a structural prior to guide the targeted detection and inpainting of lighting artifacts in RGB video. Validation on synthetic benchmarks and real-world automotive footage demonstrates that \textsc{DeLux} effectively suppresses local artifacts and restores affected regions. The proposed approach outperforms existing RGB-only baselines and event-guided HDR models, achieving an average MS-SSIM of over 0.99 across all artifact types and demonstrating up to an 88\% reduction in artifact severity in real-world automotive footage. The synthetic artifact generation tools and curated real-world evaluation datasets are made publicly available to foster future research on cross-modal restoration.
\end{abstract}

\section{Introduction}
\label{sec:intro}

\begin{figure}[hb]
    \centering
    \begin{subfigure}[t]{0.29\linewidth}
        \centering
        \includegraphics[width=1\linewidth]{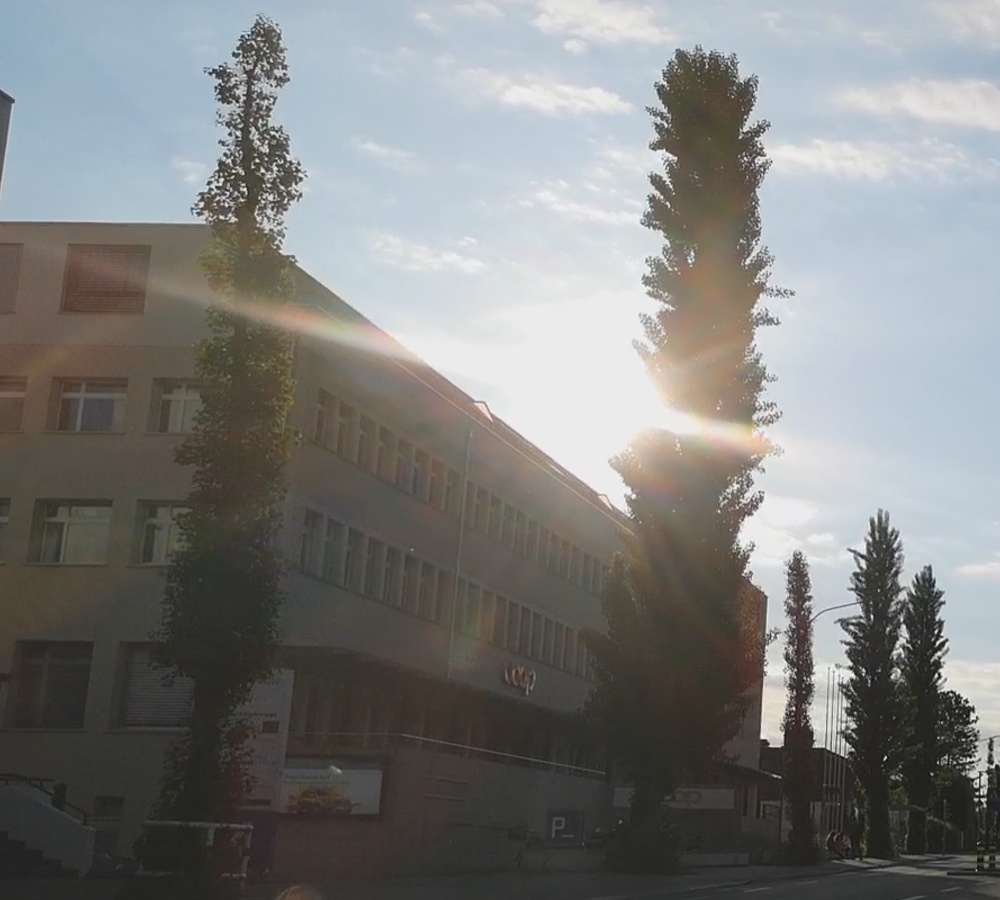}
        \caption{Glare.}
        \label{fig:glare-example}
    \end{subfigure}
    \begin{subfigure}[t]{0.29\linewidth}
        \centering
        \includegraphics[width=1\linewidth]{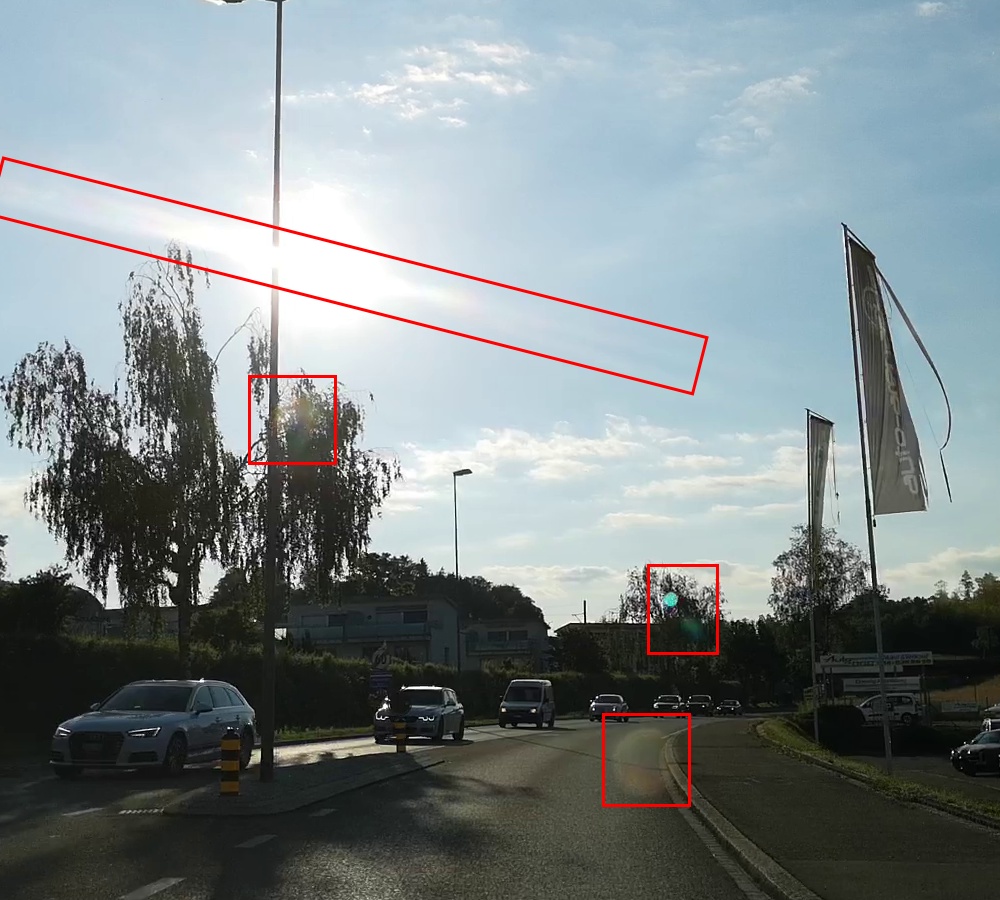}
        \caption{Lens-flare.}
        \label{fig:lens-flare-example}
    \end{subfigure}
    \begin{subfigure}[t]{0.29\linewidth}
        \centering
        \includegraphics[width=1\linewidth]{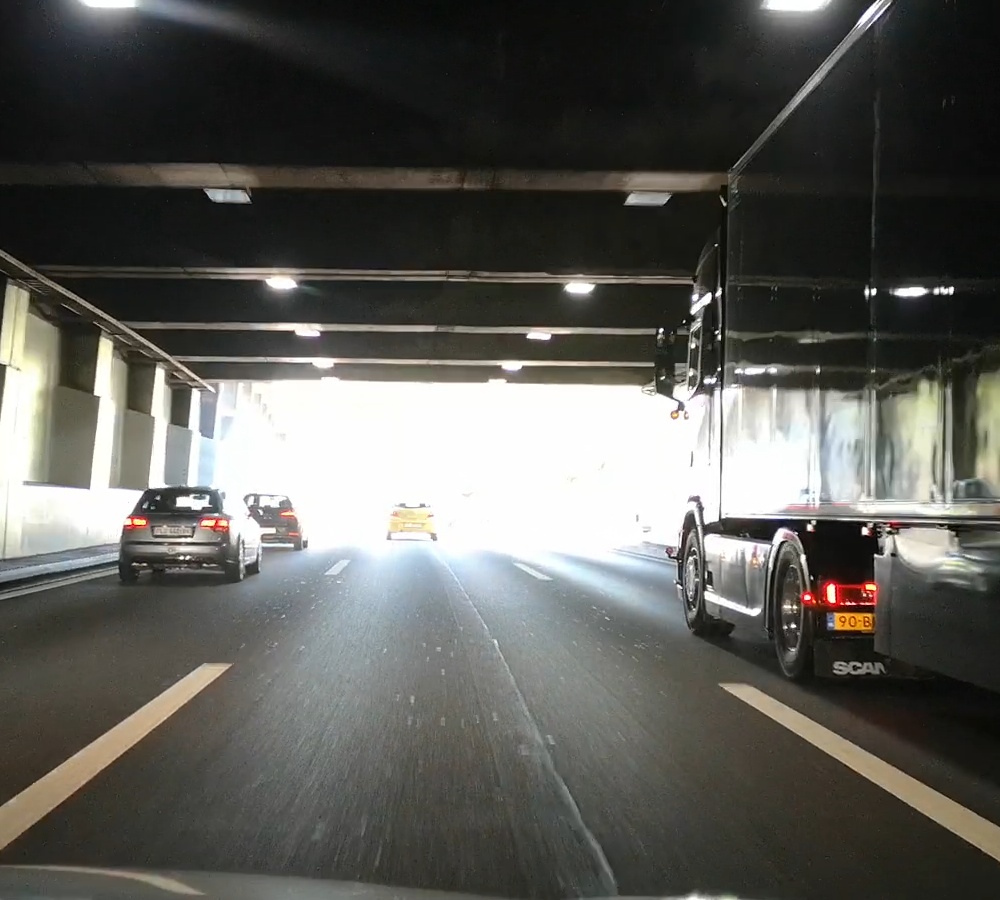}
        \caption{Overexposure.}
        \label{fig:overexposure-example}
    \end{subfigure}
    \begin{subfigure}[t]{0.9\linewidth}
        \centering
        \includegraphics[width=1\linewidth]{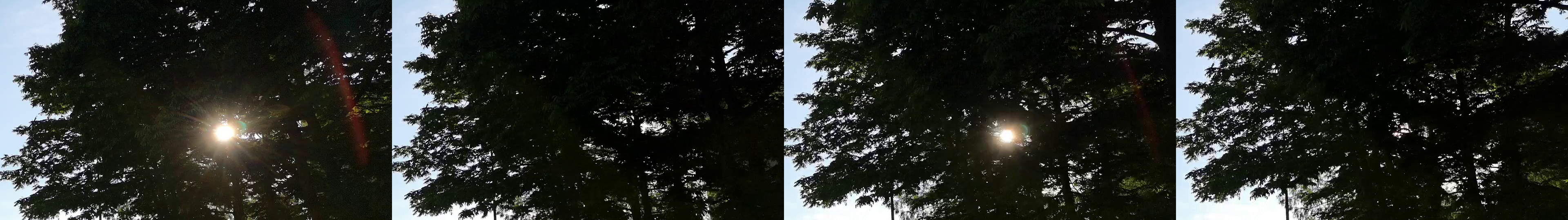}
        \caption{Flicker shown across four subsequent video frames.}
        \label{fig:flicker-example}
    \end{subfigure}
    \caption{Examples of lighting artifacts.}
    \label{fig:artifact-examples}
\end{figure}

Outdoor scenes with extreme dynamic range present a persistent challenge for conventional cameras. 
When exposed to intense or variable illumination, RGB sensors often produce \textit{ lighting artifacts} such as lens flare \cite{flare-removal}, glare \cite{cr-gan-glare-removal}, overexposure \cite{multi-image-hdr-with-dl}, and flicker \cite{removing-flicker} (Fig. \ref{fig:artifact-examples}). These artifacts obscure critical visual details, distort local color and contrast, and, in extreme cases, completely saturate the image, making recovery through RGB-only post-processing impossible.  Such degradations can severely limit the reliability of visual perception systems in applications such as autonomous driving and robotics~\cite{emergent-sensors,rgb-camera-failures}.

However, existing image and video restoration methods address these artifacts only \textit{in isolation}, for example, by mitigating only glare~\cite{cr-gan-glare-removal, veiling-glare-removal, glare-removal-in-laparoscopic-images}, flare~\cite{flare-removal, flare7kpp, nighttime-flare-removal}, global overexposure~\cite{event-cgans-hdr, deep-cnns-for-single-image-hdr, outdoor-for-single-image-hdr, shdr, yang2023learning, yang2025event, li2024generalizing} or flicker~\cite{removing-flicker, flicker-superpixel-motion, blind-deflickering}. Moreover, there is a difference in how these lighting problems are tackled. High dynamic range (HDR) reconstruction methods perform global tone mapping by compressing the dynamic range but inherently preserve the structure of optical artifacts, such as flare rays or glare halos. Conversely, targeted artifact removal methods rely solely on frame-based RGB information, which lacks the dynamic range needed to reconstruct regions where pixel intensities are fully clipped. Therefore, a main challenge in the literature is a highly fragmented approach to image degradation. Because localized optical anomalies and extreme dynamic range deficiencies are treated as disjoint sub-fields, the literature lacks a generalized paradigm capable of addressing diverse lighting artifacts simultaneously.

\begin{figure}[htb]
    \centering
    \includegraphics[width=0.95\linewidth]{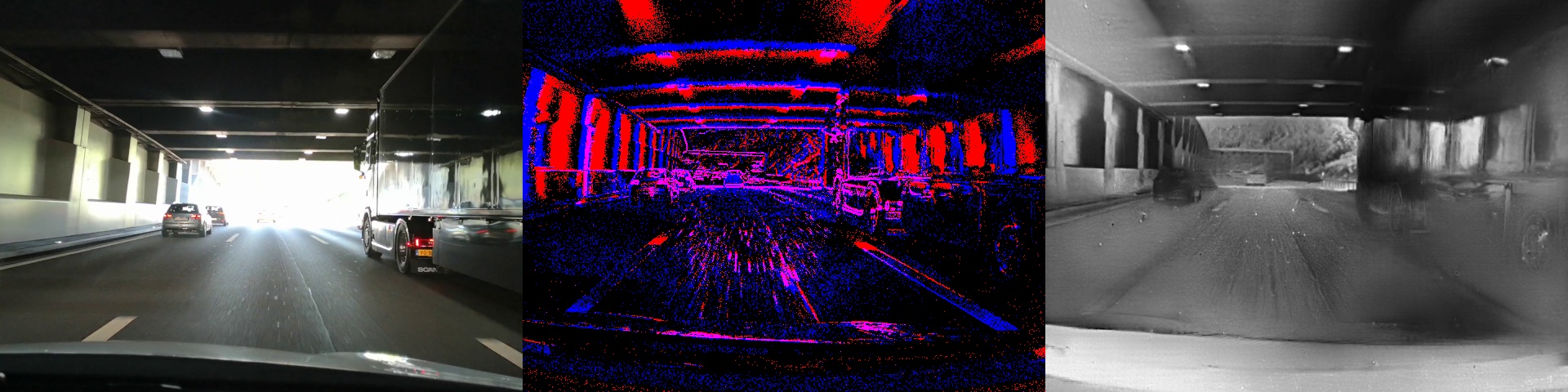}
    \caption{Comparison showing that event-based data can recover details obscured by lighting artifacts. The overexposed RGB image (left) loses all structure at the tunnel exit, while the corresponding raw event data (center) retain brightness changes. The event-based reconstruction (right) restores the scene details.}
    \label{fig:events-motivation}
\end{figure}

In this work, we propose a new research direction that combines the challenges of all the mentioned image degradations: \textit{cross-modal local artifact restoration}. In this problem setting, localized lighting artifacts represent a modality-specific information loss that can be mitigated by integrating a secondary sensing modality. To validate this concept, we use neuromorphic event cameras, which asynchronously capture changes in brightness with microsecond precision and an effective dynamic range exceeding 140\,dB \cite{event-survey}, while remaining largely unaffected by glare and saturation even when RGB sensors saturate (Fig.~\ref{fig:events-motivation}).

To further validate the feasibility of treating various lighting artifacts as instances of the proposed cross-modal local artifact restoration problem, we present \textsc{DeLux}, an end-to-end, modular proof-of-concept pipeline for lighting artifact mitigation that jointly leverages RGB video and neuromorphic event data (Fig.~\ref{fig:our-pipeline}). \textsc{DeLux} explicitly isolates the detection of lighting artifacts from the reconstruction of missing image details, using event data to guide the inpainting of corrupted RGB regions. To the best of our knowledge, \textsc{DeLux} is the first system designed to detect and remove multiple types of lighting artifacts within a single deep learning pipeline.

The key contributions of our work are as follows:
\begin{itemize}
    \item We introduce a novel research direction for targeted removal of lighting artifacts that leverages complementary sensor modalities to guide RGB inpainting, moving beyond global tone-mapping approaches and single modality image restoration.
    \item We propose \textsc{DeLux}, a modular pipeline that demonstrates the feasibility of this paradigm by fusing standard video with neuromorphic event data to jointly detect and remove diverse lighting artifacts.
    \item We establish a rigorous benchmark comprising synthetically generated and real-world lighting artifacts. We demonstrate the efficacy of our approach against state-of-the-art restoration and event-guided HDR baselines, providing an evaluation protocol for future cross-modal artifact restoration models.
\end{itemize}

\section{Related Works}
\label{sec:related-works}

\paragraph{Glare Removal.}
Glare, caused by direct light sources such as the sun or headlights, manifests itself as haze or rays that obscure the details of the scene \cite{cr-gan-glare-removal} (Fig.~\ref{fig:glare-example}). Early methods focused on reflective glare in medical imaging \cite{medical-packaging-glare-removal,glare-removal-in-laparoscopic-images}. For outdoor veiling glare, Shoshin et al. \cite{veiling-glare-removal} trained a UNet with dual branches on synthetic data to reduce haze and restore intensity, although prominent rays persisted. CR-GAN \cite{cr-gan-glare-removal} employs a GAN with a Cascaded Pyramid Neck and Glare Attention Detector, trained on synthetic and real paired data (e.g., via impure glass), achieving better perceptual quality and precise reconstruction. Furthermore, general-purpose restoration frameworks such as Deep Adversarial Decomposition (DAD) \cite{deep-adversarial-decomposition} and Uformer~\cite{uformer} offer flexible architectures that can be adapted to remove glare. DAD formulates image separation as a generative adversarial task, using a learned decomposition prior to cross-road loss to disentangle superimposed layers, which could be extended to isolate glare components. Uformer, a Transformer-based U-Net variant, leverages window-based attention and multi-scale modulation to restore fine details across diverse degradation types, making it suitable for training on glare-specific artifacts. These approaches (excluding Uformer) use artifact detectors, but focus solely on glare and rely on RGB, which limits recovery in fully saturated areas. Our work extends this by incorporating neuromorphic data for cross-modal restoration.

\paragraph{Flare Removal.}
Lens flare arises from internal reflections within the camera lens, creating ghosts or streaks (Fig.~\ref{fig:lens-flare-example}). Physical solutions such as anti-reflective coatings \cite{ar-flare-preventive-measure} are impractical for post-capture correction. Computational methods include Wu et al.'s~\cite{flare-removal} UNet pipeline, which removes flares in a semi-supervised fashion by using synthetic masks that blend backlight sources and introduce flares to the training data. Matta et al.~\cite{gn-flare-removal} use Generalizable NeRFs with a transformer to model flares in 3D scenes, supported by a multi-flare dataset. For nighttime flares, Dai et al.'s Flare7K \cite{nighttime-flare-removal} enables training of MFDNet~\cite{mfdnet-nighttime-flare-removal}, a Laplacian pyramid-based model for low-light distortions. 
We have incorporated flares from Flare7K++~\cite{flare7kpp} into our training pipeline, but contrary to the works mentioned above, we additionally use event streams and restore the underlying images rather than simply performing RGB-only inpainting.

\paragraph{Overexposure.}
Overexposure in high dynamic range (HDR) scenes saturates bright areas, hiding image details (Fig.~\ref{fig:overexposure-example}). Multi-image HDR combines LDR exposures \cite{event-multi-bracket-for-multi-image-hdr,multi-image-hdr-with-dl,multi-image-hdr-with-dl-2}, often with CNNs for alignment. Single-image methods map LDR to HDR using UNets \cite{deep-cnns-for-single-image-hdr,deep-recip-for-single-image-hdr,expandnet-for-single-image-hdr,outdoor-for-single-image-hdr} or reverse pipelines \cite{reverse-camera-pipeline-for-single-image-hdr}. SHDR \cite{shdr} explicitly models the inverse camera pipeline—dequantization, linearization, and hallucination—using dedicated CNNs, improving HDR reconstruction fidelity across diverse scenes. Related works enhance low-light~\cite{low-light-image-enhancement-for-electric-stations}, combine over/underexposed frames~\cite{combining-over-and-under-exposed-weld-frames-for-high-quality-frame}, or inpaint overexposed regions with transformers~\cite{inpainting-overexposed-fragments-for-segmentation} or GANs~\cite{aabgan-for-over-under-exposure}. BacklitNet~\cite{backlitnet-for-backlit-image-enhancement} targets overexposure with a dual-UNet. Recent methods, such as HDRev \cite{yang2023learning} and HDRev-Diff \cite{yang2025event}, use event streams to guide HDR reconstruction. However, like all other approaches that tackle overexposure, they are designed to perform global tone mapping and dynamic range expansion. Because they do not separate the underlying scene from superimposed degradations, they inherently preserve and reproduce lighting artifacts, such as lens flare and veiling glare, within the output image. In contrast, our study addresses the problem of explicit suppression of local artifacts.

\paragraph{Flicker Removal.}
Flicker is usually defined as periodic intensity variations from lights or water refraction (Fig.~\ref{fig:flicker-example}). For underwater sunflicker, Shihavuddin et al.~\cite{flicker-texture-preidction} predict patterns with an open-loop linear dynamic system from prior frames. In high-speed video, Kanj et al.~\cite{flicker-superpixel-motion} use SLIC superpixels and color correction matrices. These require tuning and are computationally heavy, focusing on compensation rather than artifact removal. Deflicker~\cite{blind-deflickering} is a blind deflickering framework that leverages a flawed neural atlas and a filtering strategy to enforce temporal consistency without requiring prior knowledge.

\paragraph{Neuromorphic Vision.}
Neuromorphic cameras capture asynchronous events with high temporal resolution and a wide dynamic range \cite{survey-on-event-representation-for-image-processing}. Event representations such as voxel grids \cite{voxel-grid-event-representation} enable integration with CNNs/Transformers for tasks such as deblurring \cite{learning-event-based-motion-deblurring,event-deblurring-ecir}, superresolution \cite{event-superresolution-gs}, depth estimation \cite{event-depth-recurrent-transformer}, and HDR reconstruction \cite{event-camera-for-hdr,event-cgans-hdr, yang2025event}. Datasets like DSEC \cite{dsec-dataset} and E2VID~\cite{e2vid} provide paired RGB event streams, while simulators like CARLA \cite{carla} generate synthetic data. We use voxel grids with E2VID++ \cite{e2vid,e2vid_plus} for artifact-free grayscale reconstruction, combining it with RGB via a ResUNet \cite{resnet}. Previous event-HDR works \cite{event-multi-bracket-for-multi-image-hdr,event-camera-for-hdr} inspire our approach, but they focus on global enhancement rather than local artifact removal. The proposed \textsc{DeLux} pipeline represents a departure from global enhancement, serving instead as a cross-modal local artifact restoration framework for lighting anomalies.

\section{DeLux Pipeline}
\label{sec:delux-pipeline}

To demonstrate the utility of the  cross-modal local artifact restoration paradigm and to address the challenge of information loss in RGB frames corrupted by lighting artifacts, we introduce \textsc{DeLux}, a pipeline that leverages event camera data to recover scene details. Unlike prior methods that rely solely on RGB input or global tone mapping, our approach uses cross-modal cues to localize and suppress artifacts at their source.

\textsc{DeLux} is designed around four core stages (Fig.~\ref{fig:our-pipeline}): (1) \textit{event-to-video reconstruction}, (2) \textit{artifact detection}, (3) \textit{multimodal fusion}, and (4) \textit{artifact removal}. Each stage is implemented as a standalone module, allowing flexible substitution or extension. This modularity enables future integration of alternative modalities, such as thermal imagery or manual annotations, thereby making \textsc{DeLux} a template for cross-modal local artifact removal.

\subsection{Event-to-video Reconstruction}

The first stage of the \textsc{DeLux} pipeline transforms neuromorphic event streams into dense grayscale video frames that serve as a complement to RGB input. These reconstructions are critical for restoring regions affected by lighting artifacts, particularly those that are saturated or flickering, where RGB data alone fail to preserve scene details.

We utilize pre-trained E2VID \cite{e2vid} and E2VID++ \cite{e2vid_plus} models that convert asynchronous event data into intensity frames. The input to this model is a voxel grid representation of events, constructed by binning spatio-temporal event tuples into a fixed number of temporal slices. Each voxel encodes the polarity and timestamp of brightness changes at a given pixel, capturing fine-grained motion and illumination dynamics. The E2VID models synthesize grayscale frames from these voxel inputs. The output frames are temporally aligned with the RGB stream and serve as a secondary modality for both detection and fusion. Additional details of the event reconstruction and spatiotemporal alignment can be found in the Supplementary, Section A.

Importantly, this stage is modular---the event-to-video reconstructor can be replaced with alternative modalities (e.g., thermal imagery \cite{thermal-not-affected-by-light}), provided that they offer complementary information in artifact-prone regions. This design choice ensures that \textsc{DeLux} remains adaptable to diverse sensor configurations and application domains.

\subsection{Artifact Detection}

The second stage of the \textsc{DeLux} pipeline identifies regions affected by lighting artifacts such as glare, flare, flicker, and overexposure. These artifacts manifest locally and vary in spatial characteristics, making their detection a non-trivial task. To address this, \textsc{DeLux} uses a dedicated artifact detector implemented as a UNet-based segmentation network.

The detector is trained on synthetically generated artifact masks, allowing it to generalize across diverse lighting conditions and artifact types. Synthetic training data includes controlled examples of saturation, flicker patterns, glare streaks, and lens flare overlays. The input to the detector is a single RGB frame, and the output is a pixel-wise confidence map that indicates the likelihood of artifact presence. These maps are used in subsequent stages to guide the image restoration, ensuring that corrections are applied only where necessary. 

Unlike prior approaches that embed artifact detection implicitly within a GAN or restoration network, \textsc{DeLux} treats detection as an explicit and interpretable step. This separation improves transparency, facilitates debugging, and enables a targeted evaluation of detection performance against synthetic ground-truth artifact masks.

\subsection{Multimodal Fusion}
The third stage of the \textsc{DeLux} pipeline integrates information from three modalities---RGB frames, secondary modality (in this case event-based reconstructions), and artifact masks---into a unified representation that guides artifact-aware restoration. This fusion is implemented as a convolutional module designed to selectively leverage the strengths of each input: the spatial richness of RGB, the temporal and dynamic range resilience of event reconstructions, and the localization precision of the artifact detector.

The three inputs are concatenated along the channel dimension and passed through a series of convolutional layers that learn to blend them into a coherent intermediate representation. This fusion strategy is fully differentiable and is trained together with the artifact removal module (Section~\ref{sec:delux-pipeline:removal}). Rather than relying on learned attention, the module blends the three channel-concatenated representations through stacked convolutions, while the soft artifact mask remains exposed as a manual-override interface. This makes fusion behavior configurable at inference \emph{without retraining}, where supplying an edited or empty mask predictably changes which regions are modified (Supplementary Section~E). By explicitly conditioning the fusion on artifact localization, \textsc{DeLux} avoids overcorrecting clean regions. This targeted blending is critical for preserving fine details and maintaining perceptual consistency (Section~\ref{sec:experiments:ablation}, Supplementary Section B).

\subsection{Artifact Removal}
\label{sec:delux-pipeline:removal}
The final stage of \textsc{DeLux} performs targeted restoration of RGB frames by removing lighting artifacts identified in earlier stages. This module receives the fused representation, comprising the original RGB frame, the event-based reconstruction, and the artifact mask, and outputs a clean, artifact-free RGB image.

Artifact removal is implemented using a UNet-based architecture trained to inpaint corrupted regions while preserving unaffected content. To encourage perceptual consistency and structural fidelity, the model is trained using a combination of pixel-wise losses, structural, and perceptual similarity metrics. In addition, the pipeline supports optional use of a non-mask loss term to regularize global appearance and prevent overfitting to masked regions.

The modular design of this stage enables adaptation to various artifact types or restoration goals. For example, the removal network can be retrained to target specific distortions such as flare streaks or flicker bands, or extended to operate on alternative input modalities. This flexibility makes \textsc{DeLux} suitable for a wider range of configurations.

\subsection{Training Procedure}
\label{sec:training}

\begin{figure*}[ht]
\centering
\begin{subfigure}[t]{0.56\textwidth}
	\centering
	\begin{minipage}[t]{0.2\textwidth}
		\vspace{1cm}
		\resizebox{\textwidth}{!}{%
		\begin{minipage}{2.5cm}
		\raggedright
		{\scriptsize \textbf{Legend:}}\\[0.1em]
		\renewcommand{\arraystretch}{0.5}
        \tiny
		\begin{tabular}{@{}l@{}}
			\texttt{\colorbox{gray!30}{CB~}} Conv Block \\
			\texttt{\colorbox{blue!20}{UCB}} UpConv Block \\
			\texttt{\colorbox{red!30}{MP~}} MaxPool 2D \\
			\texttt{\colorbox{green!30}{BN~}} 2D Batch Normaliztion \\
			\texttt{\colorbox{darkgray!30}{C~~}} ($k{\times}k$) 2D Convolution \\
			\texttt{\colorbox{cyan!30}{TC~}} ($k{\times}k$) Trans. 2D Conv. \\
			\texttt{\colorbox{orange!30}{BI~}} Bilinear Interpolation \\
		\end{tabular}
		\end{minipage}%
		}%
	\end{minipage}
	\begin{minipage}[t]{0.73\textwidth}
		\centering
		{\scriptsize \textbf{U-Net Architecture}}
		\vspace{0.3em}
		
		\resizebox{\textwidth}{!}{%
\footnotesize
\begin{tikzpicture}[ node distance=1.5cm ]
\tikzstyle{layer} = [rectangle, text width = 1.5cm, text centered, draw = black, fill = gray! 30, minimum width=1cm, minimum height = 0.5cm]
\tikzstyle{upconv} = [fill=blue! 20]
\tikzstyle{activation} = [fill = yellow! 30]
\tikzstyle{maxpool} = [fill = red! 30]
\tikzstyle{skip} = [draw=gray, thick, dashed, ->]
\tikzstyle{flow} = [->, thick]

\node (det_l1) [layer, rotate=90, minimum width=3cm] { CB };
\node (input) [left of=det_l1, text width = 1cm] {Input\\Image};
\node (det_a1) [layer, right of=det_l1, maxpool, rotate=90, yshift=1cm, minimum width=3cm] { MP };

\node (det_dot_1) [right of=det_l1] {$\dots$};
\node (det_l2) [layer, right of=det_dot_1, rotate=90, minimum width=2cm, yshift=0.5cm, xshift=-0.5cm] { CB };
\node (det_a2) [layer, right of=det_l2, maxpool, rotate=90, yshift=1.5cm, minimum width=2cm, yshift=-0.5cm] { MP };

\node (det_l3) [layer, right of=det_l2, rotate=90, minimum width=1cm, text width=1cm, xshift=-0.35cm] { CB };
\node (det_a3) [layer, right of=det_l3, maxpool, rotate=90, yshift=1cm, minimum width=1cm, text width=1cm] { MP };

\node (det_l4) [layer, upconv, right of=det_l3, rotate=90, minimum width=2cm, xshift=0.3cm] { UCB };

\node (det_dot_2) [right of=det_l4, yshift=0.5cm] {$\dots$};

\node (det_l5) [layer, upconv, right of=det_dot_2, rotate=90, minimum width=3cm, yshift=0.5cm] { UCB };

\node (det_l6) [layer,  right of=det_l5, rotate=90, minimum width=3cm, yshift=0.5cm, text width=2cm] { C($1\times1$)};
\node (det_a6) [layer, right of=det_l6, activation, rotate=90, yshift=1cm, minimum width=3cm] { Sigmoid };

\node (output) [right of=det_a6, text width=1cm] {Output\\Image};

\draw[decorate, decoration={brace, mirror, amplitude=10pt}] (-0.5,-1.55) -- (3.5,-1.55) node[midway, yshift=-20pt] {$n$ blocks};

\draw[decorate, decoration={brace, mirror, amplitude=10pt}] (5,-1.55) -- (8.5,-1.55) node[midway, yshift=-20pt] {$n$ blocks};

\draw[flow] (input) -- (det_l1);
\draw[flow] (det_a1) -- (det_dot_1);
\draw[flow] (det_dot_1) -|- (det_l2);
\draw[flow] (det_a2) -|- (det_l3);
\draw[flow] (det_a3) -|- (det_l4);
\draw[flow] (det_l4) -|- (det_dot_2);
\draw[flow] (det_dot_2) -- (det_l5);
\draw[flow] (det_l5) -- (det_l6);
\draw[flow] (det_a6) -- (output);

\draw[skip] (0,1.5) -- (0,2) -- (8,2)  node[midway, yshift=-0.7cm] {Skip Connections} -- (8,1.5);
\draw[skip] (2.5,0.5) -- (2.5,0.75) -- (5.5,0.75) -- (5.5,0.5);

\end{tikzpicture}
}%
		
		\vspace{1em}
		
		\begin{minipage}[t]{0.48\textwidth}
			\centering
			{\scriptsize \textbf{Conv Block}}
			\vspace{0.3em}
			
			\resizebox{\linewidth}{!}{%
\begin{tikzpicture}[ node distance=1.5cm ]

\Large

\tikzstyle{add} = [circle, draw=black, minimum size=0.7cm, inner sep=0pt]
\tikzstyle{skip} = [draw=gray, thick, dashed, ->]
\tikzstyle{flow} = [->, thick]
\tikzstyle{layer} = [rectangle, text width = 2cm, text centered, draw = black, fill = darkgray! 30, minimum width=2cm, minimum height = 0.5cm, rotate=90]
\tikzstyle{activation} = [fill = yellow! 30]
\tikzstyle{bn} = [fill = green! 30]
\tikzstyle{add} = [circle, draw=black, minimum size=0.7cm, inner sep=0pt]

\node (input) [text width = 1cm] {Input};

\node (conv1) [layer, below of=input] { C($k\times k$) };
\node (act1) [layer, activation, below of=input, yshift=-0.7cm] {ReLU};
\node (conv2) [layer, below of=conv1, yshift=-0.5cm] { C($k\times k$) };
\node (act2) [layer, activation, below of=conv1, yshift=-1.2cm] { ReLU };
\node (dots) [right of=conv2, xshift=0.45cm] {$\dots$};
\node (conv3) [layer, below of=dots] { C($k\times k$) };
\node (act3) [layer, activation, below of=dots, yshift=-0.7cm] { ReLU };
\node (conv_skip) [layer, left of=conv1, xshift=-1.25cm] { C($1\times 1$) };
\node (add) [add, right of=conv_skip, xshift=3cm] {$+$};
\node (bnorm) [layer, bn, below of=add] {BN};
\node (act_end) [layer, activation, below of=add, yshift=-0.7cm] {ReLU};
\node (output) [right of=act_end, text width=1cm] {Output};

\draw[decorate, decoration={brace, amplitude=10pt}] ([xshift=-15pt]conv2.east) -- ([xshift=15pt]act3.east) node[midway, yshift=18pt] {$m$ layers};

\draw [flow] (input) -- (conv1);
\draw [flow] (act1) -- (conv2);
\draw [flow] (act2) -- (dots);
\draw [flow] (dots) -- (conv3);
\draw [flow] (act3.south) -| ([yshift=-0.4cm, xshift=0.4cm]act3.south west) -| (add);
\draw [flow] (add) -- (bnorm);
\draw [flow] (act_end) -- (output);

\draw [flow] (input) |- (conv_skip);
\draw [flow] (conv_skip) -- (add);

\end{tikzpicture}

}%
		\end{minipage}\hfill
		\begin{minipage}[t]{0.48\textwidth}
			\centering
			{\scriptsize \textbf{UpConv Block}}
			\vspace{0.3em}
			
			\resizebox{\linewidth}{!}{%
\begin{tikzpicture}[ node distance=1.5cm ]
\large

\tikzstyle{add} = [circle, draw=black, minimum size=0.7cm, inner sep=0pt]
\tikzstyle{skip} = [draw=gray, thick, dashed, ->]
\tikzstyle{flow} = [->, thick]
\tikzstyle{layer} = [rectangle, text width = 2cm, text centered, draw = black, fill = gray! 30, minimum width=2cm, minimum height = 0.5cm, rotate=90]
\tikzstyle{convt} = [fill = cyan! 20]
\tikzstyle{activation} = [fill = yellow! 30]
\tikzstyle{interpolation} = [fill = orange! 30]
\tikzstyle{concat} = [circle, draw=black, minimum size=0.7cm, inner sep=0pt]

\node (skipinput) [text width = 1cm] { Skip Input };
\node (upinput) [text width = 1cm, below of=skipinput, yshift=-1.25cm] {Up Input};

  \node (tconv) [layer, convt, below of=upinput] {TC($k\times k$)};
\node (tact) [layer, activation, below of=upinput, yshift=-0.5cm] {ReLU};

\node (itp) [layer, interpolation, below of=tact] {BI};
\node (concat) [concat, label=left:Concat., right of=itp, yshift=1.5cm] {\Large $\times$};

\node (convblock) [layer, below of=concat] {CB};

\node (output) [text width = 1cm, right of=convblock] { Output };

\draw [skip] (skipinput) -| (concat);
\draw [flow] (upinput) -- (tconv);
\draw [flow] (tact) -- (itp);
\draw [flow] (itp) -| (concat);
\draw [flow] (concat) -- (convblock);
\draw [flow] (convblock) -- (output);

\end{tikzpicture}
}%
		\end{minipage}
	\end{minipage}
\end{subfigure}
\begin{subfigure}[t]{0.43\textwidth}
	\centering
	{\scriptsize \textbf{Training Pipeline}}
	\vspace{0.3em}
	
	\input{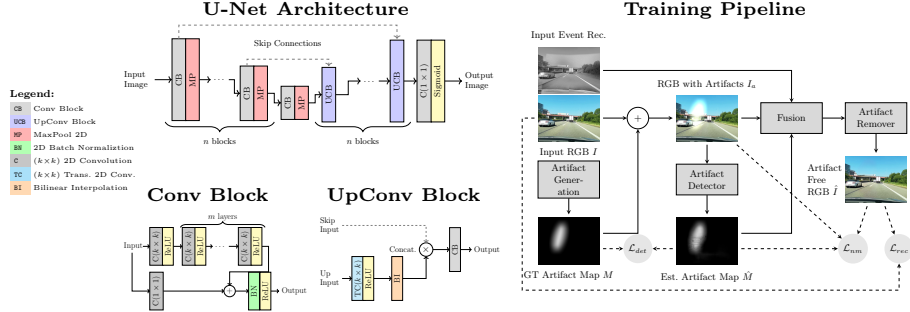}
\end{subfigure}

\caption{\textsc{DeLux} training pipeline. The architecture diagram (left) illustrates the shared U-Net backbone and its building blocks, parameterized by the number of convolutional blocks $n$, block depth $m$, and kernel size $k$.  The training pipeline (right) depicts the end-to-end joint optimization setup with synthetic artifact generation, multimodal inputs, and the composite loss functions defined in Section~\ref{sec:training}.}
\label{fig:architecture}
\end{figure*}

The \textsc{DeLux} pipeline is trained to jointly optimize the artifact detector, the multimodal fusion module, and the artifact remover (Fig.~\ref{fig:architecture}). Only the event-to-video reconstructor (E2VID or E2VID++) is frozen throughout the training, providing a stable grayscale representation of events from the neuromorphic camera. Joint optimization encourages the detector and reconstructor to co-adapt, ensuring that artifact localization and restoration are mutually consistent.

\

\paragraph{Synthetic Artifact Generation.}
To provide diverse and controlled supervision, we employ a custom artifact generation pipeline that simulates lighting distortions directly on clean RGB frames. The generators produce glare, flicker, overexposure, and saturation artifacts through intensity modulation, procedural glare streak synthesis, and localized dynamic brightness fluctuations. We additionally use high-quality, physically plausible flare patterns sourced from the Flare7K++~\cite{flare7kpp} and Google-Flare~\cite{flare-removal} datasets, ensuring a broad range of artifact shapes, scales, and illumination dynamics. Corresponding artifact masks are recorded along with the corrupted images to serve as ground truth for artifact detection. The event-based reconstructions are obtained from real neuromorphic datasets (e.g., DSEC~\cite{dsec-dataset}, E2VID~\cite{e2vid}) or generated using the CARLA simulator~\cite{carla}, providing authentic multimodal conditions. This hybrid data strategy ensures that \textsc{DeLux} generalizes effectively to real-world lighting artifacts.

\paragraph{Optimization Objective.}
The detector, fusion, and reconstructor modules are trained jointly under a unified objective that balances artifact localization and image reconstruction quality. The total loss ($\mathcal{L}_{\text{total}}$) combines three complementary components, i.e., detection ($\mathcal{L}_{\text{det}}$), reconstruction ($\mathcal{L}_{\text{recon}}$), and a non-mask consistency term ($\mathcal{L}_{\text{nm}}$):
\begin{multline}
\label{eq:l_total_final}
\mathcal{L}_{\text{total}}(I, I_a, \hat{I}, M, \hat{M}) = 
\mathcal{L}_{\text{det}}(\hat{M}, M) +
\mathcal{L}_{\text{recon}}(\hat{I}, I) + 
\mathcal{L}_{\text{nm}}(\hat{I}, I_a, \hat{M}),
\end{multline}
where $I$ is the clean ground-truth frame, $I_a$ the synthetically corrupted input frame, $\hat{I}$ the reconstructed artifact-free RGB frame, $M$ the ground-truth artifact mask, and $\hat{M}$ the predicted artifact confidence map.

The detection loss $\mathcal{L}_{\text{det}}$ is a Huber loss~\cite{huber-loss} between $\hat{M}$ and $M$, using $\delta{=}0.1$ to maintain sensitivity to small prediction errors. The reconstruction loss $\mathcal{L}_{\text{recon}}$ combines pixel-wise, perceptual, and structural similarity terms:
\begin{multline}
\mathcal{L}_{\text{recon}}(\hat{I}, I) = 
\lambda_1 \mathrm{MAE}(\hat{I}, I) + 
\lambda_2 (1 - \mathrm{MS\text{-}SSIM}(\hat{I}, I)) +
\lambda_3 \mathrm{VGG}(\hat{I}, I) \\+ 
\lambda_4 \mathrm{TV}(\hat{I}),
\end{multline}
where $\mathrm{MAE}$ encourages accurate pixel reconstruction, $\mathrm{MS\text{-}SSIM}$ promotes perceptual and structural fidelity~\cite{msssim}, $\mathrm{VGG}$ denotes the perceptual feature loss~\cite{perceptual-loss}, and $\mathrm{TV}$ regularizes spatial smoothness~\cite{tv-loss}. In this study, we manually set the coefficients to $\lambda_1{=}1.0$, $\lambda_2{=}0.5$, $\lambda_3{=}0.1$, $\lambda_4{=}0.05$.

The non-mask loss $\mathcal{L}_{\text{nm}}$ penalizes undesired edits outside artifact regions and promotes consistency in clean areas:
\begin{multline}
\mathcal{L}_{\text{nm}}(\hat{I}, I_a, \hat{M}) =
\mathrm{MAE}(\hat{I}, I_a) \odot (1 - \hat{M}) \\+
0.5 \cdot \big(1 - \mathrm{MS\text{-}SSIM}((1 - \hat{M}) \odot \hat{I}, (1 - \hat{M}) \odot I_a)\big),
\end{multline}
where $\odot$ denotes element-wise multiplication. This term guides the reconstructor to modify only artifact-corrupted regions, preserving color and texture in unaffected areas, at the cost of less accurate artifact detection maps.

\paragraph{Training Details.}
The network was trained using the Adam optimizer ($\beta_1{=}0.9$, $\beta_2{=}0.999$) with an initial learning rate of $10^{-4}$. Training was performed for $200$ epochs on an NVIDIA A100 GPU with a batch size of~$8$. Each batch contained temporally aligned RGB frames, event-based reconstructions, and artifact masks. Data augmentation included random cropping, horizontal flipping, exposure jitter, and brightness perturbation to improve robustness to illumination changes. A hold-out validation split was used to monitor convergence, and the best model was selected based on the minimum reconstruction loss  $\mathcal{L}_{\text{recon}}$. The hyperparameters of the UNet were optimized with Optuna~\cite{optuna}, resulting in $n{=}3$, $m{=}2$, $k{=}7$ for the detector and $n{=}2$, $m{=}3$, $k{=}7$ for the reconstructor (Fig.~\ref{fig:architecture}).

Training used a composite dataset combining real and synthetic event-based data for robustness across domains. Only artifact-free recordings from \textbf{E2VID}~\cite{e2vid} and \textbf{DSEC}~\cite{dsec-dataset} were used, with synthetic lighting artifacts injected during training. Additional data were generated using the \textbf{CARLA} autonomous driving simulator~\cite{carla} to provide perfectly aligned RGB--event pairs, and the \textbf{Cityscapes}~\cite{cityscapes} dataset contributed close to $5{,}000$ RGB frames to which we created synthetic grayscale reconstructions that simulate event data. All datasets were standardized to $640{\times}480$ resolution, with recordings downsampled to every third frame ($102$~ms for E2VID, $150$~ms for DSEC and CARLA) to ensure diversity in terms of motion, illumination, and scene content. The entire training set consisted of $9{,}540$ RGB-event reconstruction frame pairs.\footnote{The code and data for training \textsc{DeLux} are available at: \url{https://github.com/Tremirre/event-sun-effects-remover}.}

\section{Experiments}
\label{sec:experiments}

\subsection{Experimental Setup}

\paragraph{Datasets.}
We used two complementary groups of datasets, both distinct from all training data. The first consisted of $1
{,}724$ artifact-free frames from separate recordings, to which synthetic artifacts were injected using the same generation procedure as in training. The second comprised $11{,}056$ frames from eight real-world recordings with lighting artifacts: six from the \textbf{E2VID} dataset~\cite{e2vid} and two from \textbf{DSEC}~\cite{dsec-dataset}. Importantly, these recordings represent real-world driving scenarios exhibiting unconstrained motion and dynamic lighting. We isolated specific temporal windows in which severe lighting artifacts occur natively, establishing the first targeted evaluation subset for localized artifact removal in real-world automotive contexts.

\paragraph{Baseline Methods.}
To evaluate our \textsc{DeLux} pipeline, we compared it with a diverse set of representative baselines tailored to different artifact type. For lens flare and glare removal, we selected Flare7K++~\cite{flare7kpp} and a model developed by Wu et al.~\cite{flare-removal}. For overexposure correction, we employed SHDR~\cite{shdr}, with HDR to PNG conversion done using the algorithm outlined by Liang et al.~\cite{hdr-to-png}. For global HDR reconstruction using event data, we used HDRev-Diff~\cite{yang2025event}. Due to the scarcity of pretrained open-source models specifically designed for the analyzed artifacts, we additionally trained a variant of the universal DAD framework~\cite{deep-adversarial-decomposition} (denoted DAD\textsuperscript{\textdagger}) to separate artifacts from the underlying clean image based on the RGB and grayscale event reconstructions. Notably, the DAD variant was trained on the same data as \textsc{DeLux} and the proposed training pipeline. Therefore, DAD provides a direct comparison between an existing single neural network and the proposed modular approach.

\paragraph{Evaluation Metrics.}
The tested models were separately evaluated in terms of artifact \textit{detection} and \textit{removal}. 

For evaluating artifact detection, we used synthetic datasets where ground-truth artifact-free images were available. The detection performance was assessed using the accuracy and F1 score of the predicted artifact maps. 

The efficacy of artifact removal was quantified using both synthetic and real-world data. For the synthetic data, we computed the Multi-Scale Structural Similarity Index Measure (MS-SSIM) \cite{msssim}, Peak Signal-to-Noise Ratio (PSNR), and Mean Absolute Percentage Error (MAPE) to compare the restored final images against the corresponding clean reference frames. In contrast, real-world data lack corresponding artifact-free ground truth, necessitating the use of alternative quality assessment strategies. As such, we introduce a custom metrics: Strong Artifact Suppression (SAS, Eq.~\ref{eq:sas}). SAS depends on a reliable artifact detector, as it compares detection maps before ($\hat{M}_{input}$) and after ($\hat{M}_{output}$) processing by the model.
    \begin{equation}
    \label{eq:sas}
        \mathrm{SAS} = P(\hat{M}_{input} > 0.5) - P(\hat{M}_{output} > 0.5)
    \end{equation}
As the artifact detector, we used the \textsc{DeLux} with the non-mask loss $\mathcal{L}_{\text{nm}}$ turned off  (denoted as \textsc{DeLux-D}), as it produced the most accurate detection maps in our ablation studies (see Section~\ref{sec:experiments:ablation}). Because SAS is computed from a learned detector, we treat it as a relative artifact-suppression indicator suited to ranking methods rather than as an absolute score. To confirm that the ranking is not an artifact of using our own detector, we also computed SAS with the independently trained
DAD\textsuperscript{\textdagger} detector, obtaining similar results (Supplementary Section~B.2).

\subsection{Results for Artifact Detection}
\label{sec:experiments:detection}
Table~\ref{tab:synth-det-results} reports detection accuracy and F1 scores on synthetic artifacts. \textsc{DeLux-D} achieves the highest performance in nearly all categories, surpassing the other multimodal models, i.e., DAD\textsuperscript{\textdagger} and the full \textsc{DeLux} variant. Its advantage is most pronounced for glare, overexposure, and complex flares. Disabling the non-mask loss term $\mathcal{L}_{\text{nm}}$ in \textsc{DeLux-D} improves artifact map details because the artifact detection is `disconnected' from image reconstruction. 
However, as we will be shown in Section~\ref{sec:experiments:ablation}, the stronger and slightly larger estimated artifact maps of the full \textsc{DeLux} variant offer better reconstruction quality. 
The flare-specific models (Flare7K++ and Wu et al.) fail to generalize beyond their training domains. A comparison of example artifact maps predicted by the analyzed models is presented in Fig.~\ref{fig:synth-artifact-detection-removal} and in the Supplementary, Section C.

\begin{table}[t]
\centering
\caption{Detection results on the synthetic test set across different artifact types. Best values are highlighted in \colorbox{orange!30}{\textbf{bold}}, second-best values are \colorbox{yellow!30}{\underline{underlined}}.}
\label{tab:synth-det-results}
\begin{NiceTabular}{llcccc|c}
\toprule
 Metric & Category & DAD\textsuperscript{\textdagger} \cite{deep-adversarial-decomposition} & F7K \cite{flare7kpp} & Wu et al. \cite{flare-removal} & DeLux & DeLux-D \\
\midrule
\multirow[c]{6}{*}{Accuracy $\uparrow$} & Glare & \cellcolor{yellow!30}\underline{0.969} & 0.914 & 0.830 & 0.964 & \cellcolor{orange!30}\textbf{0.985} \\
 & HQ Flares & \cellcolor{yellow!30}\underline{0.938} & 0.920 & 0.779 & 0.891 & \cellcolor{orange!30}\textbf{0.959} \\
 & Overexposure & \cellcolor{yellow!30}\underline{0.972} & 0.862 & 0.819 & 0.943 & \cellcolor{orange!30}\textbf{0.982} \\
 & Simple Flares & 0.973 & \cellcolor{orange!30}\textbf{0.991} & 0.880 & 0.977 & \cellcolor{yellow!30}\underline{0.989} \\
 & Flares+Glare & 0.936 & 0.835 & 0.799 & \cellcolor{yellow!30}\underline{0.939} & \cellcolor{orange!30}\textbf{0.973} \\
\cline{2-7}
 & Overall & \cellcolor{yellow!30}\underline{0.949} & 0.915 & 0.809 & 0.925 & \cellcolor{orange!30}\textbf{0.971} \\
\cline{1-7}
\multirow[c]{6}{*}{F1-score $\uparrow$} & Glare & 0.805 & 0.145 & 0.339 & \cellcolor{yellow!30}\underline{0.815} & \cellcolor{orange!30}\textbf{0.900} \\
 & HQ Flares & \cellcolor{yellow!30}\underline{0.601} & 0.292 & 0.301 & 0.530 & \cellcolor{orange!30}\textbf{0.739} \\
 & Overexposure & \cellcolor{yellow!30}\underline{0.793} & 0.188 & 0.409 & 0.719 & \cellcolor{orange!30}\textbf{0.841} \\
 & Simple Flares & 0.484 & 0.065 & 0.045 & \cellcolor{yellow!30}\underline{0.574} & \cellcolor{orange!30}\textbf{0.749} \\
 & Flares+Glare & 0.744 & 0.191 & 0.403 & \cellcolor{yellow!30}\underline{0.745} & \cellcolor{orange!30}\textbf{0.871} \\
\cline{2-7}
 & Overall & \cellcolor{yellow!30}\underline{0.634} & 0.212 & 0.281 & 0.613 & \cellcolor{orange!30}\textbf{0.785} \\
\bottomrule
\end{NiceTabular}

\end{table}

\begin{figure*}[ht]
    \centering
    \input{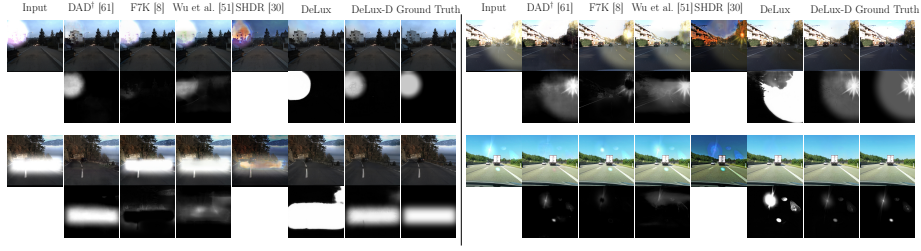}
    \caption{Qualitative comparison of artifact removal (top row) and artifact detection (bottom row) on different kinds of synthetic artifacts.}
    \label{fig:synth-artifact-detection-removal}
\end{figure*}

\subsection{Results for Artifact Removal}
Table~\ref{tab:full-real-results} summarizes the results of artifact removal. On synthetic data with available ground truth, \textsc{DeLux} achieves the highest MS-SSIM, PSNR, and MAPE across all artifact types, confirming the feasibility and strong restoration fidelity of the proposed cross-modal inpainting approach. 

\begin{table}[!t]
\scriptsize
\centering
\caption{Results of artifact removal evaluation. Best values are highlighted in \colorbox{orange!30}{\textbf{bold}}, second-best are \colorbox{yellow!30}{\underline{underlined}}. Rows labeled `Overall' denote the score calculated using combined data from all categories. The $\Delta$ symbol signifies the relative metric change compared to the input image. HDRev-Diff was not evaluated on reconstruction metrics as the examples from Cityscapes used synthetic grayscale reconstructions instead of event streams.}
\label{tab:full-real-results}
\begin{NiceTabular}{l@{\quad}ll@{}r@{\,\,\,}r@{\,\,\,}r@{\,\,\,}r@{\,\,\,}r@{\,\,\,}r}
\toprule
 & Metric & Category & DAD\textsuperscript{\textdagger}  & F7K  & Wu et al.  & SHDR  & HDRev-Diff & DeLux \\
 &        &          & \cite{deep-adversarial-decomposition} & \cite{flare7kpp} & \cite{flare-removal} & \cite{shdr} & \cite{yang2025event} & (Ours) \\
\midrule
\multirow[c]{18}{*}{\rotatebox[origin=c]{90}{\centering Ground Truth Reconstruction}} & \multirow[c]{6}{*}{MS-SSIM $\uparrow$} & Glare & \cellcolor{yellow!30}\underline{0.969} & 0.938 & 0.938 & 0.859 & - & \cellcolor{orange!30}\textbf{0.990} \\
 &  & High-Quality Flares & \cellcolor{yellow!30}\underline{0.968} & 0.957 & 0.950 & 0.867 & - & \cellcolor{orange!30}\textbf{0.990} \\
 &  & Overexposures & \cellcolor{yellow!30}\underline{0.959} & 0.891 & 0.872 & 0.813 & - & \cellcolor{orange!30}\textbf{0.981} \\
 &  & Simple Flares & \cellcolor{yellow!30}\underline{0.984} & 0.979 & 0.978 & 0.894 & - & \cellcolor{orange!30}\textbf{0.997} \\
 &  & Sun Flares+Glare & \cellcolor{yellow!30}\underline{0.967} & 0.949 & 0.947 & 0.868 & - & \cellcolor{orange!30}\textbf{0.990} \\
\cline{3-9}
 &  & Overall & \cellcolor{yellow!30}\underline{0.970} & 0.954 & 0.949 & 0.868 & - & \cellcolor{orange!30}\textbf{0.991} \\
\cline{2-9}
 & \multirow[c]{6}{*}{PSNR $\uparrow$} & Glare & \cellcolor{yellow!30}\underline{28.279} & 20.753 & 21.195 & 13.522 & - & \cellcolor{orange!30}\textbf{35.287} \\
 &  & High-Quality Flares & \cellcolor{yellow!30}\underline{29.567} & 27.888 & 26.767 & 14.688 & - & \cellcolor{orange!30}\textbf{35.858} \\
 &  & Overexposure & \cellcolor{yellow!30}\underline{29.208} & 21.002 & 19.877 & 12.864 & - & \cellcolor{orange!30}\textbf{34.501}  \\
 &  & Simple Flares & 32.128 & \cellcolor{yellow!30}\underline{32.231} & 31.032 & 14.796 & - & \cellcolor{orange!30}\textbf{41.402}  \\
 &  & Sun Flares+Glare & \cellcolor{yellow!30}\underline{27.384} & 21.840 & 22.178 & 14.249 & - & \cellcolor{orange!30}\textbf{34.122} \\
\cline{3-9}
 & & Overall & \cellcolor{yellow!30}\underline{29.525} & 26.583 & 25.848 & 14.418 & - & \cellcolor{orange!30}\textbf{36.461} \\
\cline{2-9}
 & \multirow[c]{6}{*}{MAPE $\downarrow$} & Glare & \cellcolor{yellow!30}\underline{0.059} & 0.097 & 0.099 & 0.576 & - & \cellcolor{orange!30}\textbf{0.025} \\
 & & High-Quality Flares & \cellcolor{yellow!30}\underline{0.063} & 0.064 & 0.072 & 0.553 & - & \cellcolor{orange!30}\textbf{0.032} \\
 & & Overexposure & \cellcolor{yellow!30}\underline{0.061} & 0.135 & 0.153 & 0.531 & - & \cellcolor{orange!30}\textbf{0.031} \\
 & & Simple Flares & 0.039 & \cellcolor{yellow!30}\underline{0.025} & 0.031 & 0.442 & - & \cellcolor{orange!30}\textbf{0.015} \\
 & & Sun Flares+Glare & \cellcolor{yellow!30}\underline{0.076} & 0.107 & 0.111 & 0.576 & - & \cellcolor{orange!30}\textbf{0.034} \\
\cline{2-9}
 & & Overall & \cellcolor{yellow!30}\underline{0.060} & 0.071 & 0.078 & 0.537 & - & \cellcolor{orange!30}\textbf{0.028} \\
\midrule
\multirow[c]{9}{*}{\rotatebox[origin=c]{90}{\centering Artifact Removal}} & \multirow[c]{9}{*}{$\Delta$ SAS $\uparrow$} & Sun 02 & \cellcolor{orange!30}\textbf{87.64\%} & 78.78\% & 81.22\% & 55.46\% & 84.78\% & \cellcolor{yellow!30}\underline{87.36\%} \\
 & & Sun 05 & 87.50\% & 85.06\% & 85.61\% & -90.44\% & \cellcolor{yellow!30}\underline{87.79\%} & \cellcolor{orange!30}\textbf{87.90\%} \\
 & & Sun 06 & -130.65\% & \cellcolor{yellow!30}\underline{33.62\%} & 26.24\% & -847.04\% & \cellcolor{orange!30}\textbf{34.75\%} & -5.64\% \\
 & & Sun 09 & 88.22\% & 83.81\% & 87.62\% & 61.36\% & \cellcolor{orange!30}\textbf{89.18\%} & \cellcolor{yellow!30}\underline{88.32\%} \\
 & & Sun 10 & \cellcolor{yellow!30}\underline{87.74\%} & 81.62\% & 83.40\% & 51.76\% & 26.22\% & \cellcolor{orange!30}\textbf{87.79\%} \\
 & & Sun 11 & -173.61\% & \cellcolor{yellow!30}\underline{43.53\%} & 40.43\% & -761.38\% & -5.76\% & \cellcolor{orange!30}\textbf{51.57\%} \\
 & & Zurich City 01-e & -1953.29\% & 45.60\% & -123.38\% & -216.95\% & -7986.77\% & \cellcolor{orange!30}\textbf{66.17\%} \\
 & & Zurich City 12-a & 30.41\% & 26.00\% & -109.16\% & 51.18\% & -15.34\% & \cellcolor{orange!30}\textbf{63.83\%} \\
\cline{2-9}
 & & Overall & -352.50\% & \cellcolor{yellow!30}\underline{58.87\%} & 18.54\% & -255.65\% & -1402.87\% & \cellcolor{orange!30}\textbf{63.22\%} \\
\bottomrule
\end{NiceTabular}
\end{table}

\begin{figure*}
    \centering
    \input{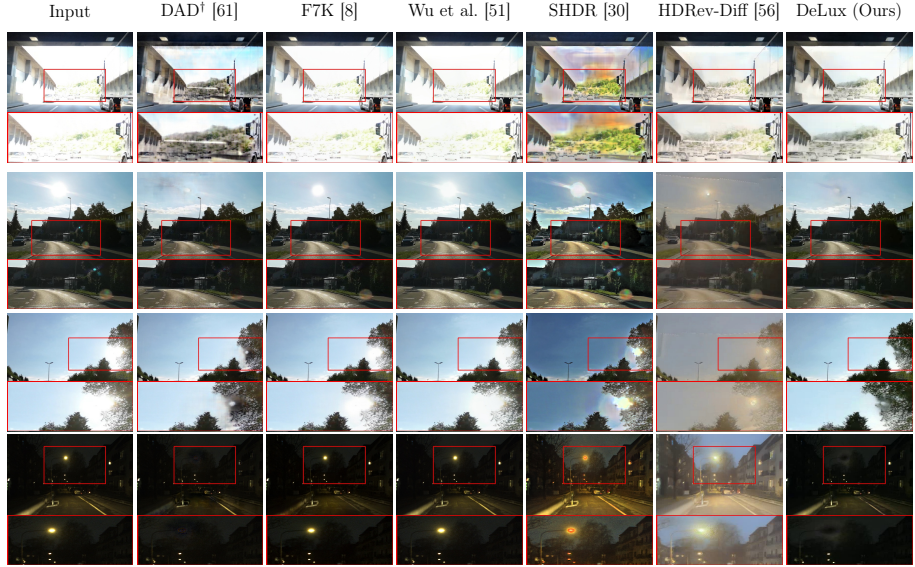}
    \caption{Comparison of artifact removal on real-world recordings. Zoom-ins (red border) provided for better visual assessment.}
    \label{fig:real-artifact-removal}
\end{figure*}

In real-world automotive scenarios, \textsc{DeLux} achieves the highest average artifact suppression (SAS), outperforming the single-network DAD\textsuperscript{\textdagger} approach in removing both mild and severe distortions. Moreover, the event-guided HDR baseline, HDRev-Diff, performed mostly global tone mapping, which, for certain scenarios, was not enough to fully suppress localized artifacts (Fig.~\ref{fig:real-artifact-removal} rows 2--4), resulting in negative SAS scores. This empirical result confirms our hypothesis: while global tone-mapping models successfully recover dynamic range, they often preserve local optical artifacts. In contrast, the explicit detection and fusion stages of \textsc{DeLux} successfully isolated and removed these degradations by using the event stream to reconstruct the underlying image structure.

Qualitative visual assessments of synthetic and real-world artifacts (Fig.~\ref{fig:synth-artifact-detection-removal} and Fig.~\ref{fig:real-artifact-removal}, and in Sections C and D of the Supplementary) further demonstrate that \textsc{DeLux} robustly handles dynamic light variations and does the best job of restoring lost image details. Overall, \textsc{DeLux} offers the most consistent balance between artifact removal and image quality.

In terms of efficiency, \textsc{DeLux} runs at 16.5\,FPS with 90.2\,M parameters on an NVIDIA A100---an order of magnitude lighter and faster than the diffusion-based HDRev-Diff baseline (1.6\,B parameters, 1.1\,FPS); full model sizes and timings are reported in the Supplementary, Section~F.

\subsection{Ablation Studies}
\label{sec:experiments:ablation}
To evaluate the impact of each architectural component of the default \textsc{DeLux} configuration, we conducted an ablation study with four variants: \textsc{DeLux-NE} (no events), \textsc{DeLux-ND} (no detector), and \textsc{DeLux-D} (disabled non-mask loss). The results are summarized in Table~\ref{tab:ablation-results}.

\begin{table}[htb]
\centering
\caption{\footnotesize Ablation study results on DeLux and its variants. Best values are \colorbox{orange!30}{\textbf{bolded}}, and second-best values are \colorbox{yellow!30}{\underline{underlined}}.}
\label{tab:ablation-results}
\begin{tabular}{lrrrrr}
\toprule
 & Accuracy $\uparrow$ & MS-SSIM $\uparrow$ & PSNR $\uparrow$ & MAPE $\downarrow$ & $\Delta$ SAS [\%] $\uparrow$ \\
\midrule
DeLux-NE & \cellcolor{yellow!30}\underline{0.964} & \cellcolor{yellow!30}\underline{0.986} & 33.757 & 0.039 & \cellcolor{orange!30}\textbf{67.06\%} \\
DeLux-ND & - & 0.982 & 30.822 & 0.059 & -195.34\% \\
DeLux-D & \cellcolor{orange!30}\textbf{0.971} &  \cellcolor{orange!30}\textbf{0.991} & \cellcolor{yellow!30}\underline{35.839} & \cellcolor{yellow!30}\underline{0.029} & -1703.38\% \\
DeLux & 0.925 & \cellcolor{orange!30}\textbf{0.991} & \cellcolor{orange!30}\textbf{36.461} & \cellcolor{orange!30}\textbf{0.028} & \cellcolor{yellow!30}\underline{63.22\%} \\
\bottomrule
\end{tabular}
\end{table}

Disabling the non-mask loss (\textsc{DeLux-D}) yields the highest artifact detection accuracy (0.971), but at the cost of artifact removal ($\Delta$ SAS of -1703.38\%). Similarly, removing the detector (\textsc{DeLux-ND}) causes the model to fail to reliably suppress artifacts ($\Delta$ SAS of -195.34\%), underscoring the need for explicit artifact localization. On the other hand, removing the event modality (\textsc{DeLux-NE}) yields the strongest apparent artifact suppression ($\Delta$ SAS of 67.06\%), but it also results in the loss of structural cues, leading to inaccurate reconstructions (PSNR=33.757).

For general-purpose scenarios requiring a balance between artifact detection and removal, we recommend the default \textsc{DeLux} variant. For strictly artifact detection tasks, we recommend the \textsc{DeLux-D} variant, which provides a detailed map of lighting artifacts. This contrast reflects a deliberate detection--removal trade-off. The non-mask loss $\mathcal{L}_{\text{nm}}$ couples the detector and inpainter, forcing edits to remain inside the predicted mask. This markedly improves removal but pushes the detector toward larger, smoother masks at the cost of pixel-sharp localization. Disabling non-mask loss (\textsc{DeLux-D}) decouples the two, yielding the sharpest masks but an unconstrained inpainter that over-modifies clean regions. The two configurations therefore target different uses---\textsc{DeLux-D} for explicit artifact maps (analysis, manual editing) and \textsc{DeLux} for end-to-end restoration.

\section{Discussion}
\label{sec:conclusions}
In this work, we introduced \textsc{DeLux}, a proof-of-concept framework that puts forward a novel research direction: cross-modal local artifact restoration. By focusing on targeted, modality-guided inpainting, we demonstrated that secondary sensing modalities, such as neuromorphic event cameras, can successfully guide the restoration of degraded RGB pixels. Our core insight is that explicitly decoupling artifact localization from multimodal fusion prevents the unwarranted alteration of clean image regions, allowing the network to selectively use the secondary modality only where the RGB sensor has failed.

However, it is also important to acknowledge certain limitations of the current study. While event cameras offer high dynamic range, they are inherently limited in stationary scenes where no events are triggered. Nevertheless, the modular architecture of \textsc{DeLux} is designed to be modality-agnostic. A possible solution to this static-scene limitation could be to combine multiple sensing modalities, for instance, by simultaneously fusing RGB, neuromorphic events, and thermal imaging. A second limitation of our current study is the use of synthetic artifact generation for evaluation against a ground-truth image. While we use high-quality flare and glare overlays (such as those from the Flare7K++ dataset), synthetic overlays cannot fully reproduce the complex physical interplay of light within a real lens assembly. The real-world evaluations, on the other hand, lacked an artifact-free equivalent. Therefore, to advance this field, future research must focus on capturing paired, real-world datasets of lighting artifacts. Furthermore, relying on a frozen E2VID reconstructor currently limits full end-to-end optimization, suggesting that future pipelines could directly ingest raw event representations to enable more efficient, task-specific feature learning. Finally, because \textsc{DeLux} restores each frame independently, its outputs are not guaranteed to be temporally consistent, and residual artifacts can vary across adjacent frames, resulting in visible flicker in the reconstructed video. Enforcing temporal consistency, e.g., through multi-frame modeling, is an important direction for future work.

Ultimately, \textsc{DeLux} serves as a template for addressing severe localized image degradations using a secondary sensing modality. By demonstrating the usefulness of cross-modal local artifact restoration, we hope to inspire research that moves beyond single-sensor constraints and paves the way for highly reliable image processing systems capable of operating under the most adverse lighting conditions.

\subsubsection*{Data and Code Availability.}
The curated real-world evaluation data, synthetic artifact generation tools, and training code are publicly available at: \url{https://github.com/Tremirre/event-sun-effects-remover}.

\bibliographystyle{splncs04}
\bibliography{main}

\renewcommand\thesection{\Alph{section}}
\renewcommand\thesubsection{\thesection.\arabic{subsection}}
\renewcommand\thefigure{S\arabic{figure}}    
\renewcommand\thetable{S\arabic{table}}  
\setcounter{section}{0}
\setcounter{figure}{0}
\setcounter{table}{0}

\section{Data Preparation Details}
\label{sec:supplement-data}

This section outlines the preparation of the training data for \textsc{DeLux}, covering the steps taken from raw recordings to the processed inputs used by the model. It also expands on the event-to-video reconstruction procedure and how event data was aligned, normalized, and integrated with RGB frames.

\subsection{Event Reconstruction}

\begin{figure}[htb]
\centering
\input{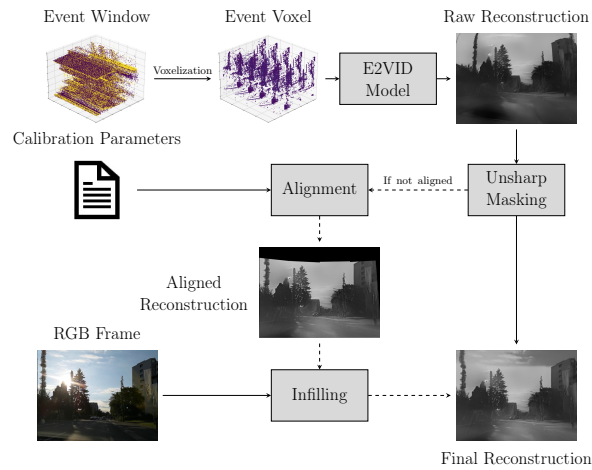}
\caption{Event-to-video reconstruction pipeline. Converts an event window into a single grayscale frame. Spatial alignment and infilling are applied only when calibration or calibration-free alignment is required.}
\label{fig:event-to-video}
\end{figure}

Figure~\ref{fig:event-to-video} presents the full event-to-video process used in our pipeline. Although this stage is built on the E2VID~\cite{e2vid} and E2VID++~\cite{e2vid_plus} models, additional processing is required to ensure that the reconstructed grayscale frames are spatially and temporally compatible with the RGB stream.

\begin{figure}[htb]
    \centering
    \begin{subfigure}[t]{0.465\linewidth}
        \centering
        \includegraphics[width=1\linewidth]{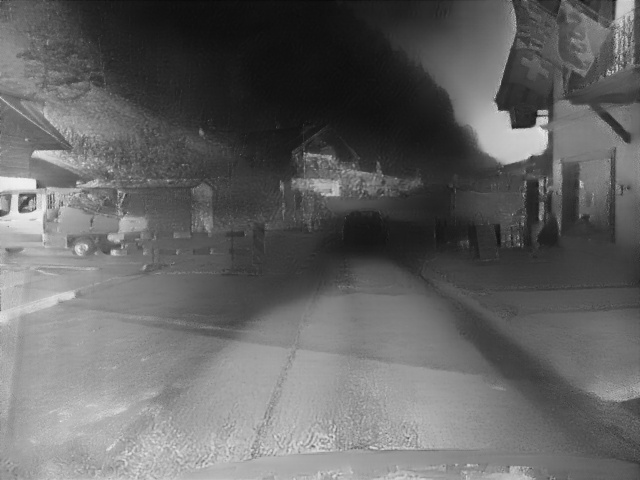}
        \captionsetup{width=0.95\linewidth}
        \caption{Reconstruction from a DSEC window using E2VID. Large regions lack meaningful detail.}
        \label{fig:interlaken-e2vid}
    \end{subfigure}
    \begin{subfigure}[t]{0.465\linewidth}
        \centering
        \includegraphics[width=1\linewidth]{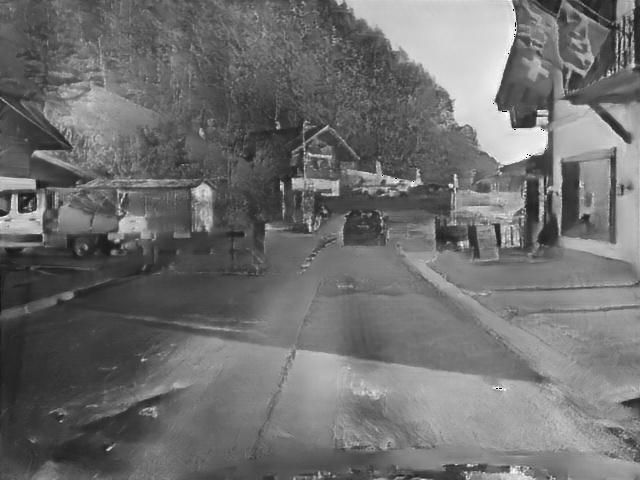}
        \captionsetup{width=0.95\linewidth}
        \caption{Reconstruction from the same window using E2VID++. Sharper and with fewer artifacts.}
        \label{fig:interlaken-e2vid-pp}
    \end{subfigure}
    \caption{Comparison of reconstructions from two event-to-video models on the same DSEC event window.}
    \label{fig:dsec-event-to-video-models-comparison}
\end{figure}

Because event-to-video models are trained on data from particular camera types, their performance varies across datasets. As shown in Figure~\ref{fig:dsec-event-to-video-models-comparison}, E2VID and E2VID++ can produce substantially different reconstructions from the same event window. For DSEC~\cite{dsec-dataset} and CARLA~\cite{carla}, E2VID++ yielded consistently sharper results, while standard E2VID was sufficient for all other datasets and served as our default choice.

\paragraph{Unsharp Masking for Detail Enhancement.}
Before any alignment or infilling is applied, the reconstructed frame undergoes unsharp masking to enhance the visibility of fine structures. This post-processing step sharpens local detail by subtracting a Gaussian-blurred version of the frame from the original and blending the residual to amplify high-frequency components~\cite{unsharp-masking}. Applying this procedure before spatial calibration ensures that structural cues—often attenuated during event-to-video reconstruction—are preserved as strongly as possible before subsequent geometric transformations.

\paragraph{Calibration and Spatial Alignment.}
When intrinsic and extrinsic calibration parameters are available, we apply them to align each reconstructed event frame with its corresponding RGB frame. Intrinsics include focal lengths, principal points, and distortion coefficients, which allow us to undistort both modalities into a common rectified image space. Extrinsics define the rigid transformation between the event and RGB cameras, enabling us to project the reconstructed event frame into the RGB coordinate system.

Applying these calibration steps may leave parts of the reconstructed event frame undefined, typically near image boundaries or in regions where the projection maps outside the event camera's field of view. To ensure complete coverage, we interpolate the missing pixels using the grayscale version of the RGB frame. This preserves spatial continuity and guarantees that subsequent stages of the pipeline operate on well-defined inputs.

\paragraph{Datasets Without Calibration.}
Some datasets, such as E2VID sequences, do not provide intrinsic or extrinsic parameters. In these cases, we rely on a calibration-free spatio-temporal alignment procedure tailored for scenarios where camera parameters are unavailable. A full description of this alignment method is provided in Section~\ref{sec:spatiotemporal-alignment} of the Supplement.

\subsection{Dataset selection}

For each used data source, we provide preparation details to facilitate better reproducibility of our results. Table~\ref{tab:data-split-counts} summarizes the composition of each subset in our hold-out validation strategy. We note that we have decided not to use all the recordings from the datasets, as the sheer volume of data posed a significant computational challenge, making it infeasible to process every recording within a reasonable time frame. Second, some recordings exhibited poor reconstruction quality, either due to sensor issues or low event density, which could compromise the integrity of the evaluation. As a result, only a curated subset of recordings was selected to balance diversity, data quality, and computational feasibility. Table~\ref{tab:dsec-e2vid-split} summarizes how we divided the recordings into separate subsets.

\begin{table}[htbp]
\centering
\caption{Number of samples (RGB frame + event reconstruction) used for training, validation, and testing with synthetic artifacts. Note that synthetic datasets (CARLA, Cityscapes) are used exclusively for training.}
\label{tab:data-split-counts}
\begin{tabular}{lrrrr}
\toprule
\textbf{Source} & \textbf{Train} & \textbf{Validation} & \textbf{Test} & \textbf{Total} \\
\midrule
CARLA         & 848   & 0     & 0    & 848   \\
Cityscapes    & 4927  & 0     & 0    & 4927  \\
DSEC          & 2425  & 925   & 940  & 4290  \\
E2VID         & 1340  & 1091  & 784  & 3215  \\
\midrule
\textbf{Total} & 9540 & 2016  & 1724 & 13280 \\
\bottomrule
\end{tabular}
\end{table}

\begin{table}[htbp]
\centering
\caption{Per-recording data split for DSEC and E2VID datasets; includes only clear recordings without real lighting artifacts.}
\label{tab:dsec-e2vid-split}
\begin{tabular}{lcccc}
\toprule
\textbf{Dataset} & \textbf{Train} & \textbf{Validation} & \textbf{Test} \\
\midrule
\textbf{DSEC} &
\begin{tabular}[c]{@{}l@{}}
interlaken\_00\_c \\
interlaken\_00\_f \\
zurich\_00\_b \\
zurich\_city\_04\_a \\
zurich\_city\_05\_b \\
zurich\_city\_06\_a
\end{tabular} &
\begin{tabular}[c]{@{}l@{}}
interlaken\_00\_g \\
zurich\_city\_07\_a
\end{tabular} &
\begin{tabular}[c]{@{}l@{}}
interlaken\_00\_d \\
zurich\_01\_a \\
zurich\_city\_13\_b
\end{tabular} \\
\midrule
\textbf{E2VID} &
\begin{tabular}[c]{@{}l@{}}
back3 \\
back4 \\
highway2
\end{tabular} &
\begin{tabular}[c]{@{}l@{}}
back1 \\
back9 \\
street2
\end{tabular} &
\begin{tabular}[c]{@{}l@{}}
highway1
\end{tabular} \\
\bottomrule
\end{tabular}
\end{table}

To avoid potential label noise and evaluation bias, we manually reviewed frames selected for supervised training via synthetic artifact injection on real-world clean backgrounds and excluded any that exhibited visible lighting artifacts. This ensures that all training, validation, and test samples with synthetic artifacts are paired with clean, artifact-free reference frames. Additionally, selected frames with real lighting artifacts that were excluded from the clean reference set were curated into a separate, manually annotated dataset used for training and evaluating the artifact detection component.

\begin{figure}[htb]
    \centering
    \begin{subfigure}[t]{0.465\linewidth}
        \centering
        \includegraphics[width=1\linewidth]{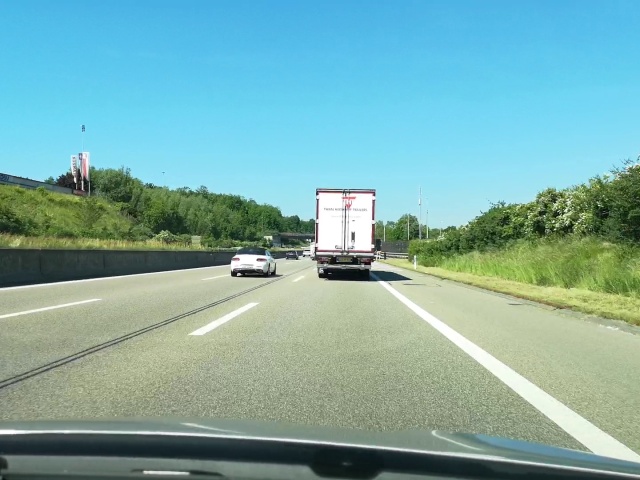}
    \end{subfigure}
    \begin{subfigure}[t]{0.465\linewidth}
        \centering
        \includegraphics[width=1\linewidth]{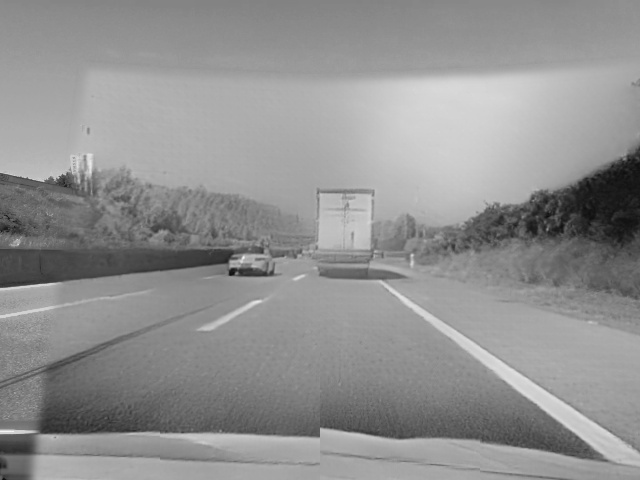}
    \end{subfigure}
    \caption{Preprocessed data sample from E2VID dataset with RGB frame data (left) and event reconstruction (right).}
    \label{fig:e2vid-data-sample}
\end{figure}

\paragraph{E2VID.}
The E2VID recordings lack intrinsic and extrinsic calibration as well as timestamp alignment. We therefore apply the calibration-free spatio-temporal alignment procedure described in Supplementary Section~\ref{sec:spatiotemporal-alignment}. Imperfect alignment can produce spatial gaps between the reconstructed event frame and the RGB frame; these are later filled during the event-to-video stage. A representative example is shown in Figure~\ref{fig:e2vid-data-sample}.

\begin{figure}[htb]
    \centering
    \begin{subfigure}[t]{0.32\linewidth}
        \centering
        \includegraphics[width=1\linewidth]{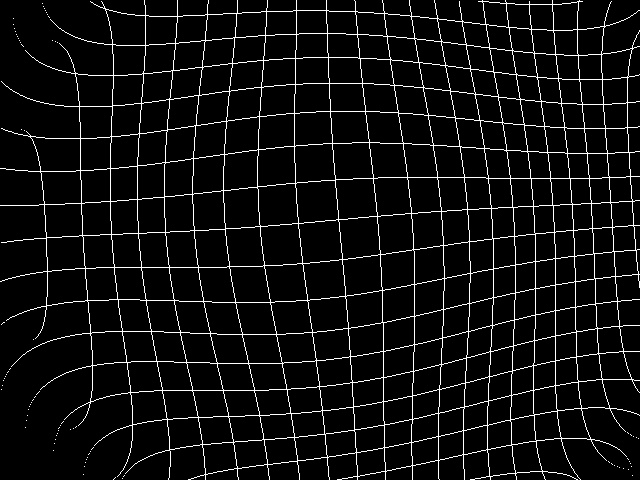}
        \captionsetup{width=0.9\linewidth}
        \caption{Binary grid depicting missing event pixels resulting from forward rectification and rounding. Rectification occurs before reconstruction, on the voxelized events.}
        \label{fig:dsec-rectify-artifact-mask}
    \end{subfigure}
    \begin{subfigure}[t]{0.32\linewidth}
        \centering
        \includegraphics[width=1\linewidth]{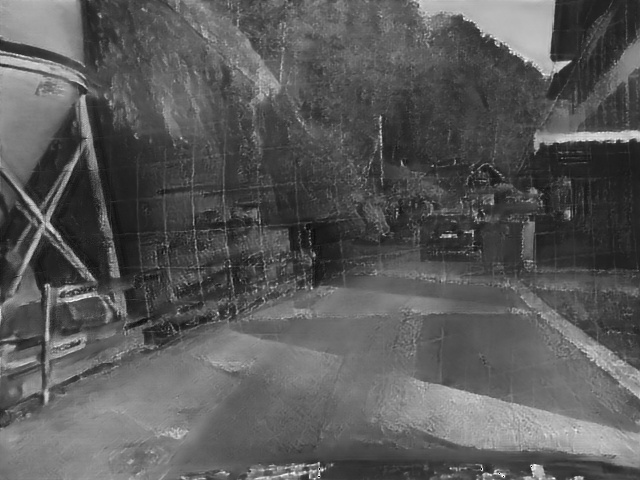}
        \captionsetup{width=0.9\linewidth}
        \caption{Reconstructed frame from one of the DSEC recordings. Model partially fills in missing information based on previous frames, but some grid outline is still visible.}
        \label{fig:dsec-reconstruction-with-grid}
    \end{subfigure}
    \begin{subfigure}[t]{0.32\linewidth}
        \centering
        \includegraphics[width=1\linewidth]{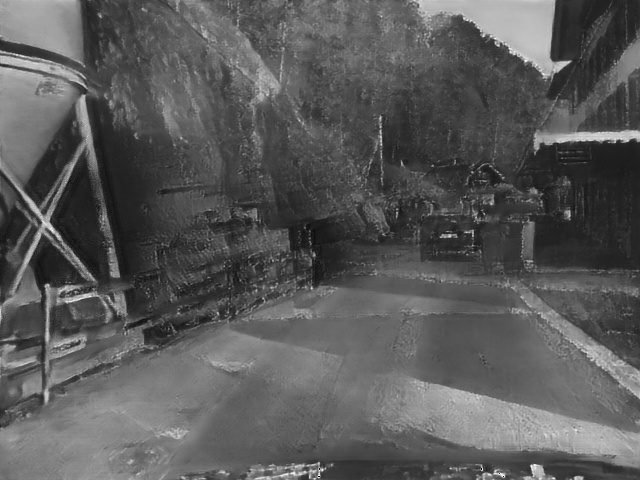}
        \captionsetup{width=0.9\linewidth}
        \caption{The same frame after replacing the pixels belonging to grid mask with average of their neighbors. Grid outline is no longer present.}
        \label{fig:dsec-reconstruction-without-grid}
    \end{subfigure}
    \caption{Grid of missing events resulting from forward rectification (A), reconstructed image before (B) and after averaging (C).}
    \label{fig:dsec-grid-removal}
\end{figure}

\begin{figure}[htb]
    \centering
    \begin{subfigure}[t]{0.465\linewidth}
        \centering
        \includegraphics[width=1\linewidth]{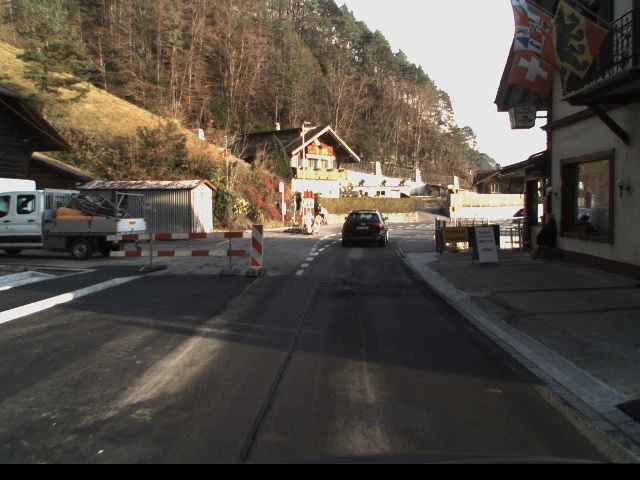}
    \end{subfigure}
    \begin{subfigure}[t]{0.465\linewidth}
        \centering
        \includegraphics[width=1\linewidth]{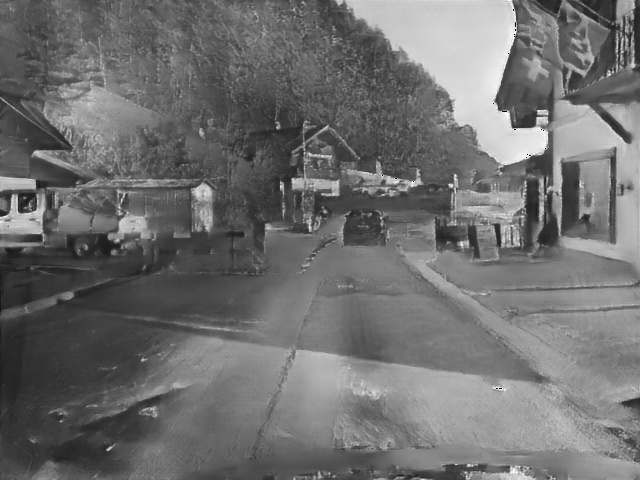}
    \end{subfigure}
    \caption{Example of spatially aligned data from the DSEC dataset:  clean RGB frame (left) and its corresponding grayscale reconstruction (right) derived from neuromorphic event data.}
    \label{fig:dsec-data-sample}
\end{figure}

\paragraph{DSEC.}
DSEC provides complete calibration, including rectification maps that project event pixels directly into the RGB frame. We apply this rectification before event reconstruction and use E2VID++ due to its higher fidelity on this dataset. Forward rectification creates a regular grid of pixels that do not receive events; after reconstruction, these pixels are filled using neighborhood averaging (Figure~\ref{fig:dsec-grid-removal}). An example aligned pair is shown in Figure~\ref{fig:dsec-data-sample}.

\begin{figure}[htb]
    \centering
    \begin{subfigure}[t]{0.465\linewidth}
        \centering
        \includegraphics[width=1\linewidth]{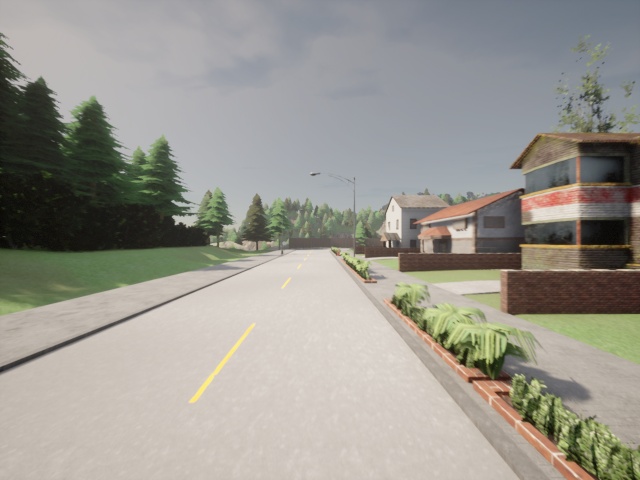}
    \end{subfigure}
    \begin{subfigure}[t]{0.465\linewidth}
        \centering
        \includegraphics[width=1\linewidth]{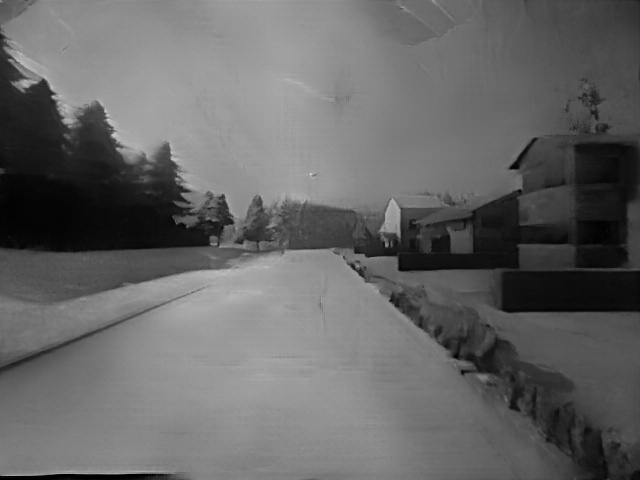}
    \end{subfigure}
    \caption{Sample frame from the CARLA simulator illustrating simulated RGB data (left) and its corresponding event-based reconstruction from simulated neuromorphic event data (right). These synthetic pairs were used to augment training data under controlled artifact-free lighting conditions.}
    \label{fig:carla-example}
\end{figure}

\paragraph{CARLA.}
To increase scene diversity and maintain full control over sensor configuration, we generate synthetic RGB and event streams in CARLA. The simulator provides perfect spatial and temporal alignment, removing the need for post-hoc calibration. Grayscale reconstructions are produced using E2VID. Although the synthetic appearance differs from the real-world imagery, these samples broaden the distribution of geometries and motion patterns (Figure~\ref{fig:carla-example}).

\begin{figure}[htb]
    \centering
    \begin{subfigure}[t]{0.465\linewidth}
        \centering
        \includegraphics[width=1\linewidth]{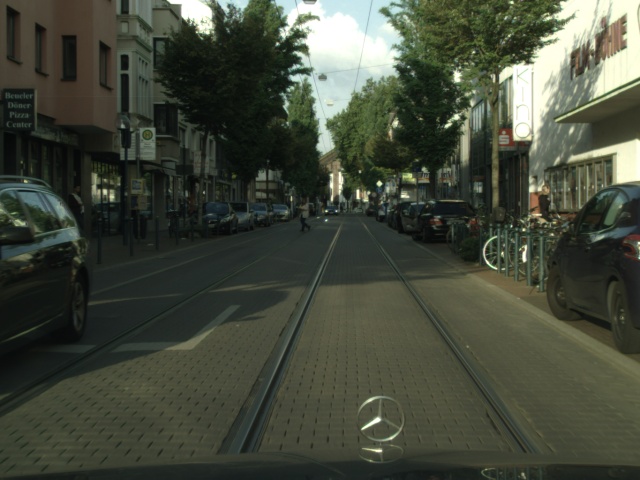}
    \end{subfigure}
    \begin{subfigure}[t]{0.465\linewidth}
        \centering
        \includegraphics[width=1\linewidth]{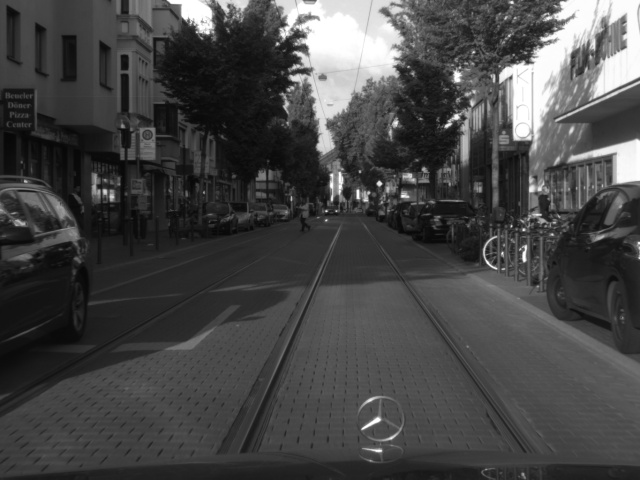}
    \end{subfigure}
    \caption{Sample RGB frame (left) from the Cityscapes dataset and its corresponding grayscale approximation (right) used to simulate event-based reconstructions in the absence of actual neuromorphic data.}
    \label{fig:cityscapes-example}
\end{figure}

\paragraph{Cityscapes.}
Cityscapes contributes high-quality urban scenes but contains no event data. We therefore use the RGB images directly and generate grayscale approximations as placeholders for event-derived reconstructions (Figure~\ref{fig:cityscapes-example}). These samples enrich the texture variety without altering the model architecture.

\subsection{Algorithm for Spatio-temporal alignment}
\label{sec:spatiotemporal-alignment}

While the \textsc{DeLux} pipeline supports aligned RGB and event data streams -- either pre-aligned or accompanied by intrinsic/extrinsic calibration parameters and a known time offset -- not all experimental setups provide this information. To address this limitation, we introduce a complementary spatio-temporal alignment procedure designed to accurately align asynchronous RGB and event data streams.

Our alignment method enables \textsc{DeLux} to be used in real-world scenarios where calibration data are missing or incomplete. It consists of three stages:

\begin{enumerate}
\item Manual Intrinsic Parameter Estimation
\item Automatic Temporal Alignment
\item Automatic Spatial Alignment
\end{enumerate}

\paragraph{Intrinsic Parameter Estimation.}

The internal camera parameters define the internal geometry of the camera, including the focal length, the principal point, and the distortion of the lens \cite{camera-calibration-geometry}. These parameters are necessary for projecting event data into the RGB image plane and are a prerequisite for accurate spatial alignment between the two modalities.

\begin{figure}[htbp]
    \centering
    \begin{subfigure}[t]{0.45\linewidth}
        \centering
        \includegraphics[width=1\linewidth]{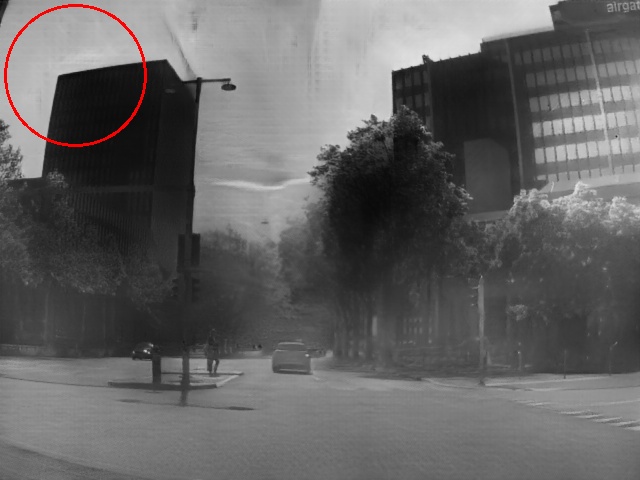}
        \captionsetup{width=0.9\linewidth}
        \caption{Reconstructed frame from one of the E2VID recordings. Visible fisheye distortion on the outer section of the frame.}
        \label{fig:distoted-example}
    \end{subfigure}
    \begin{subfigure}[t]{0.45\linewidth}
        \centering
        \includegraphics[width=1\linewidth]{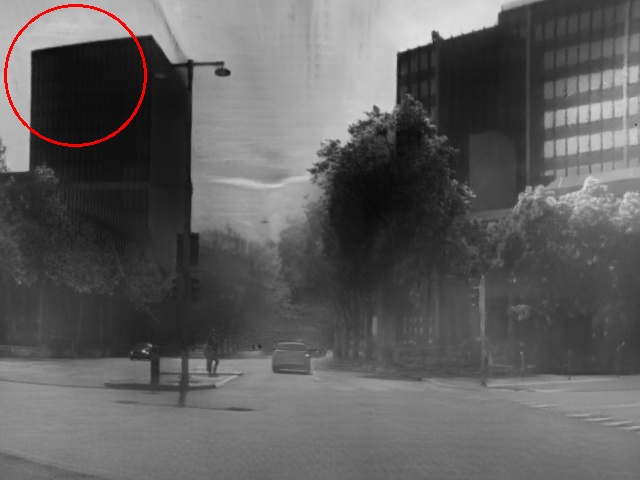}
        \captionsetup{width=0.9\linewidth}
        \caption{The same frame after correction using the estimated intrinsic parameters.}
        \label{fig:undistorted-example}
    \end{subfigure}
    \caption{Example reconstructed frame before and after distortion. Highlighted region with clear distortion before (a) and after (b) calibration.}
    \label{fig:intrinsics-calibration-result}
\end{figure}

In cases where intrinsic parameters for the event camera are not provided, we estimate them manually by overlaying straight lines onto a reconstructed event frame and iteratively adjusting the intrinsic values until the projected lines align with known linear features in the scene. An exemplary result of intrinsic calibration is presented in Figure \ref{fig:intrinsics-calibration-result}.

\paragraph{Temporal Alignment.}
Temporal alignment refers to the synchronization of the timestamps of the RGB frames with those of the event data stream. Due to differences in sensor hardware and recording pipelines, these two modalities often exhibit a time offset, even if they are nominally recorded simultaneously. Accurate temporal alignment is crucial to ensure that the reconstructed event frames correspond to the correct RGB frames.

\begin{figure}[htbp]
    \centering
    \includegraphics[width=0.7\linewidth]{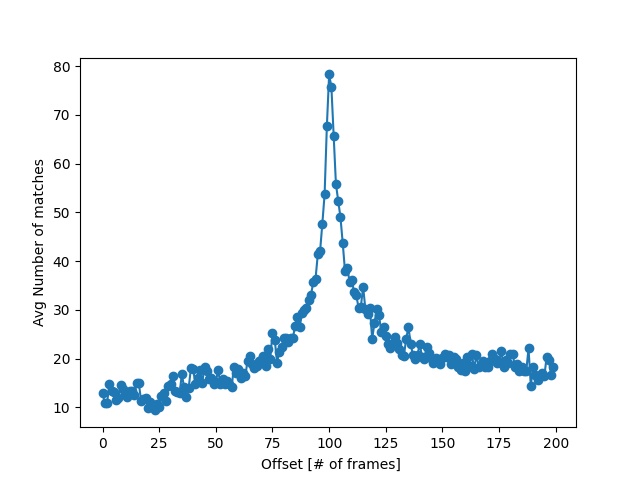}
    \caption{Chart depicting average number of matches using the SIFT feature matching algorithm \cite{sift-feature-matching} per frame offset. For each offset, the features are matched between the reconstructed frames and frames from the RGB camera, assuming this offset. The offset containing the highest average number of matches is selected as the correct one. This method assumes that the correct offset is within the tested offsets.}
    \label{fig:temporal-alignment-chart}
\end{figure}

To address this, we propose an algorithm that estimates the temporal offset between two video sequences, without requiring them to be spatially aligned (Algorithm \ref{alg:temporal_alignment}). As a prerequisite, the event stream needs to be converted to discrete frames using an event-to-video model. These reconstructed frames are then compared against RGB frames using feature matching at multiple potential offsets. The offset that yields the highest number of consistent matches is selected as optimal, as depicted in Figure \ref{fig:temporal-alignment-chart}. The algorithm is agnostic to the choice of feature extraction and matching methods, for which a comprehensive overview is provided by Jakubovic \textit{et al.}~\cite{feature-matching-overviews}.

This approach relies on the assumption that correct alignment yields a stronger correspondence between visual features in both modalities. By testing a range of offsets and scoring them by average feature match count, we find the offset that best synchronizes the data streams.

\begin{algorithm}[htbp]
\scriptsize
\caption{\scriptsize Temporal Alignment of RGB Video and Event Data}
\label{alg:temporal_alignment}
\begin{algorithmic}[1]
\Require RGB video $V$, event data $E$, window length $w$, number of offsets to check $n_o$, frame check interval $c_i$
\Ensure Temporal offset $t_{\mathit{offset}}$, aligned frame pairs $A_{best}$

\Function{TemporalAlignment}{$V, E, w, n_o, c_i$}
    \State $V \gets$ ResizeVideo($V$, $E$.width, $E$.height)
    \State $F_E \gets$ ReconstructFrames($E$, $w$) \Comment{Using E2VID neural network}
    
    \State $M \gets \emptyset$ \Comment{Array to store match counts for each offset}
    \State $A \gets \emptyset$ \Comment{Array to store frame correspondences}
    
    \For{$i \gets 0$ \textbf{to} $n_o - 1$}
        \State $M_i \gets \emptyset$ \Comment{Matches for current offset}
        \State $A_i \gets \emptyset$ \Comment{Aligned indices for current offset}
        \For{$j \gets 0$ \textbf{to} $|V|$ \textbf{step} $c_i$}
            \State $t_{aligned} \gets i \cdot w + V[j].timestamp$
            \State $k \gets \arg\min_k |F_E[k].timestamp - t_{aligned}|$
            \State $A_i \gets A_i \cup \{(j, k)\}$
            \State $f_V \gets$ ExtractFeatures($V[j]$)
            \State $f_E \gets$ ExtractFeatures($F_E[k]$)
            \State $m \gets$ CountMatches($f_V$, $f_E$)
            \State $M_i \gets M_i \cup \{m\}$
        \EndFor
        \State $M \gets M \cup \{\text{avg}(M_i)\}$
        \State $A \gets A \cup \{A_i\}$
    \EndFor
    
    \State $t_{idx} \gets \arg\max_i M[i]$
    \State $t_{\mathit{offset}} \gets t_{idx} \cdot w$ \Comment{Convert to milliseconds}
    \State $A_{best} \gets A[t_{idx}]$
    
    \State \Return $t_{\mathit{offset}}$, $A_{best}$
\EndFunction
\end{algorithmic}
\end{algorithm}

\begin{figure}[htb]
    \centering
    \includegraphics[width=1\linewidth]{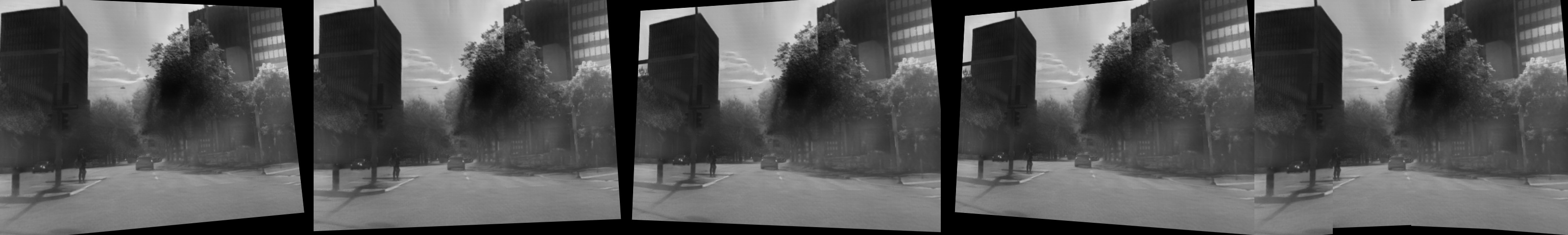}
    \caption{Four fully transformed frames using strip-wise homographies, followed by the final composite frame obtained by joining the corresponding transformed regions.}
    \label{fig:homography-alignment-result}
\end{figure}

\clearpage
\paragraph{Spatial Alignment.}
Following temporal synchronization, spatial alignment (Algorithm \ref{alg:spatial_alignment}) computes geometric transformations between the corresponding event reconstructions and RGB frames. To accommodate complex distortions that might be present in the image plane, we divide each frame into vertical strips and calculate separate homography transformations for each region (as shown in Figure \ref{fig:homography-alignment-result}). This multi-homography approach effectively handles spatial variations that a single global transform could not represent adequately. The exact number of vertical strips is a hyperparameter that can be adjusted on a per-recording basis.

The output of this stage consists of a set of per-strip homography matrices, along with the previously estimated temporal offset. Together, these parameters enable near pixel-level alignment between the RGB and event data streams, ensuring consistency in downstream tasks such as fusion, reconstruction, and artifact removal.

\begin{algorithm}[!h]
\scriptsize
\caption{\scriptsize Spatial Alignment of Temporally Aligned RGB-Event Frames}
\label{alg:spatial_alignment}
\begin{algorithmic}[1]
\Require RGB video $V$, reconstructed event frames $F_E$, aligned frame pairs $A$, number of homography strips $n_h$
\Ensure Homography transforms $\{H_1, H_2, \ldots, H_{n_h}\}$

\Function{SpatialAlignment}{$V, F_E, A, n_h$}
    \State $S \gets \emptyset$ \Comment{Array of strip specifications}
    
    \For{$(src\_idx, rec\_idx) \in A$}
        \State $masks \gets$ CreateHorizontalMasks($V[src\_idx]$.shape, $n_h$)
        \State $S_{curr} \gets \emptyset$
        
        \For{$mask \in masks$}
            \State $f_V \gets$ ExtractFeatures($V[src\_idx]$, $mask$)
            \State $f_E \gets$ ExtractFeatures($F_E[rec\_idx]$, $mask$)
            \State $matches \gets$ MatchFeatures($f_V$, $f_E$)
            \State $H \gets$ ComputeHomography($matches$)
            \State $S_{curr} \gets S_{curr} \cup \{(mask, H, |matches|)\}$
        \EndFor
        
        \State $S \gets S \cup \{S_{curr}\}$
    \EndFor
    
    \State $best\_idx \gets \arg\max_i \sum_{j=1}^{|S[i]|} S[i][j].matches$ \Comment{Best alignment maximizes sum of matches}
    \State $best\_specs \gets S[best\_idx]$
    \State $H_{transforms} \gets \{H_1, H_2, \ldots, H_{n_h}\}$ from $best\_specs$
     \State \Return $H_{transforms}$
\EndFunction
\end{algorithmic}
\end{algorithm}

\section{Ablation Studies}
\label{sec:supplement-ablation}

\subsection{Component Ablations}

In this section, we provide extended results from ablation studies on the \textsc{DeLux} pipeline. Table~\ref{tab:synth-ab-sdet-results} aggregates synthetic detection results, whilst Table~\ref{tab:ab-full-real-results} aggregates synthetic and real artifact removal results. The visual comparison for synthetic frames is shown in Figure~\ref{fig:ab-synth-artifact-detection-removal} and for real frames in Figure~\ref{fig:ab-real-artifact-removal}.

\begin{table}[htbp]
\footnotesize
\centering
\caption{Reconstruction results for ablated models on the synthetic test set across different flare types and overexposures. Best values are \colorbox{orange!30}{\textbf{bolded}}, and second-best values are \colorbox{yellow!30}{\underline{underlined}}.}
\label{tab:synth-ab-sdet-results}
\begin{NiceTabular}{llcccc}
\toprule
 Metric & Category & DeLux & DeLux-D & DeLux-ND & DeLux-NE \\
\midrule
\multirow[c]{6}{*}{Acc $\uparrow$} & Flares+Glare & 0.939 & \cellcolor{orange!30}\textbf{0.973} & - & \cellcolor{yellow!30}\underline{0.968} \\
 & Glare & 0.964 & \cellcolor{orange!30}\textbf{0.985} & - & \cellcolor{yellow!30}\underline{0.981} \\
 & HQ Flares & 0.891 & \cellcolor{orange!30}\textbf{0.959} & - & \cellcolor{yellow!30}\underline{0.951} \\
 & Overexposure & 0.943 & \cellcolor{orange!30}\textbf{0.982} & - & \cellcolor{yellow!30}\underline{0.976} \\
 & Simple Flares & 0.977 & \cellcolor{orange!30}\textbf{0.989} & - & \cellcolor{yellow!30}\underline{0.983} \\
\cline{2-6}
 & Overall & 0.925 & \cellcolor{orange!30}\textbf{0.971} & - & \cellcolor{yellow!30}\underline{0.964} \\
\cline{1-6}
\multirow[c]{6}{*}{F1 $\uparrow$} & Flares+Glare & 0.745 & \cellcolor{orange!30}\textbf{0.871} & - & \cellcolor{yellow!30}\underline{0.842} \\
 & Glare & 0.815 & \cellcolor{orange!30}\textbf{0.900} & - & \cellcolor{yellow!30}\underline{0.879} \\
 & HQ Flares & 0.530 & \cellcolor{orange!30}\textbf{0.739} & - & \cellcolor{yellow!30}\underline{0.626} \\
 & Overexposure & 0.719 & \cellcolor{orange!30}\textbf{0.841} & - & \cellcolor{yellow!30}\underline{0.820} \\
 & Simple Flares & \cellcolor{yellow!30}\underline{0.574} & \cellcolor{orange!30}\textbf{0.749} & - & 0.453 \\
\cline{2-6}
 & Overall & 0.613 & \cellcolor{orange!30}\textbf{0.785} & - & \cellcolor{yellow!30}\underline{0.666} \\
\bottomrule
\end{NiceTabular}
\end{table}

\begin{table}[htbp]
\centering
\caption{Results across all evaluation metrics and datasets. Best values are \colorbox{orange!30}{\textbf{bolded}}, and second-best values are \colorbox{yellow!30}{\underline{underlined}}.}
\label{tab:ab-full-real-results}
\begin{NiceTabular}{l@{\quad}llrrrr}
\toprule
 & Metric & Category & DeLux & DeLux-D & DeLux-ND & DeLux-NE \\
\midrule
 \multirow[c]{18}{*}{\rotatebox[origin=c]{90}{\centering Ground Truth Reconstruction}} &\multirow[c]{6}{*}{MS-SSIM $\uparrow$} & Flares+Glare & \cellcolor{orange!30}\textbf{0.990} & \cellcolor{orange!30}\textbf{0.990} & 0.980 & \cellcolor{yellow!30}\underline{0.985} \\
 &  & Glare & \cellcolor{orange!30}\textbf{0.990} & \cellcolor{orange!30}\textbf{0.990} & 0.980 & \cellcolor{yellow!30}\underline{0.982} \\
 &  & HQ Flares & \cellcolor{orange!30}\textbf{0.990} & \cellcolor{orange!30}\textbf{0.990} & 0.979 & \cellcolor{yellow!30}\underline{0.986} \\
 &  & Overexposure & \cellcolor{orange!30}\textbf{0.981} & \cellcolor{orange!30}\textbf{0.981} & \cellcolor{yellow!30}\underline{0.971} & 0.962 \\
 &  & Simple Flares & \cellcolor{orange!30}\textbf{0.997} & \cellcolor{orange!30}\textbf{0.997} & 0.992 & \cellcolor{yellow!30}\underline{0.995} \\
 \cline{3-7}
 &  & Overall & \cellcolor{orange!30}\textbf{0.991} & \cellcolor{orange!30}\textbf{0.991} & 0.982 & \cellcolor{yellow!30}\underline{0.986} \\
 \cline{2-7}
 & \multirow[c]{6}{*}{PSNR $\uparrow$} & Flares+Glare & \cellcolor{orange!30}\textbf{34.122} & \cellcolor{yellow!30}\underline{33.779} & 29.428 & 31.723 \\
 &  & Glare & \cellcolor{orange!30}\textbf{35.287} & \cellcolor{yellow!30}\underline{34.543} & 29.999 & 32.05 \\
 &  & HQ Flares & \cellcolor{orange!30}\textbf{35.858} & \cellcolor{yellow!30}\underline{35.312} & 30.474 & 33.641 \\
 &  & Overexposure & \cellcolor{orange!30}\textbf{34.501} & \cellcolor{yellow!30}\underline{34.282} & 30.174 & 31.206 \\
 &  & Simple Flares & \cellcolor{orange!30}\textbf{41.402} & \cellcolor{yellow!30}\underline{40.262} & 33.636 & 37.593 \\
 \cline{3-7}
 &  & Overall & \cellcolor{orange!30}\textbf{36.461} & \cellcolor{yellow!30}\underline{35.839} & 30.822 & 33.757 \\
 \cline{2-7}
 & \multirow[c]{6}{*}{MAPE $\downarrow$} & Flares+Glare & \cellcolor{orange!30}\textbf{0.034} & \cellcolor{yellow!30}\underline{0.036} & 0.067 & 0.046 \\
 &  & Glare & \cellcolor{orange!30}\textbf{0.025} & \cellcolor{yellow!30}\underline{0.028} & 0.061 & 0.04 \\
 &  & HQ Flares & \cellcolor{orange!30}\textbf{0.032} & \cellcolor{yellow!30}\underline{0.032} & 0.062 & 0.041 \\
 &  & Overexposure & \cellcolor{orange!30}\textbf{0.031} & \cellcolor{yellow!30}\underline{0.032} & 0.059 & 0.052 \\
 &  & Simple Flares & \cellcolor{orange!30}\textbf{0.015} & \cellcolor{yellow!30}\underline{0.017} & 0.041 & 0.024 \\
 \cline{3-7}
 &  & Overall & \cellcolor{orange!30}\textbf{0.028} & \cellcolor{yellow!30}\underline{0.029} & 0.059 & 0.039 \\
\midrule
\multirow[c]{9}{*}{\rotatebox[origin=c]{90}{\centering Artifact Removal}} & \multirow[c]{9}{*}{$\Delta$ SAS $\uparrow$} & Sun 02 & 87.36\% & \cellcolor{orange!30}\textbf{89.36\%} & 88.37\% & 88.21\% \\
 &  & Sun 05 & \cellcolor{orange!30}\textbf{87.90\%} & \cellcolor{yellow!30}\underline{87.78\%} & 86.14\% & 66.94\% \\
 &  & Sun 06 & \cellcolor{yellow!30}\underline{-5.64\%} & -10298.50\% & -1183.78\% & \cellcolor{orange!30}\textbf{34.31\%} \\
 &  & Sun 09 & 88.32\% & \cellcolor{orange!30}\textbf{89.87\%} & 87.68\% & \cellcolor{yellow!30}\underline{89.46\%} \\
 &  & Sun 10 & \cellcolor{orange!30}\textbf{87.79\%} & 76.39\% & \cellcolor{yellow!30}\underline{87.38\%} & 86.75\% \\
 &  & Sun 11 & \cellcolor{orange!30}\textbf{51.57\%} & -1594.92\% & \cellcolor{yellow!30}\underline{51.39\%} & 50.93\% \\
 &  & Zurich City 01-e & \cellcolor{yellow!30}\underline{66.17\%} & \cellcolor{orange!30}\textbf{78.96\%} & -401.58\% & 63.67\% \\
 &  & Zurich City 12-a & 63.83\% & \cellcolor{yellow!30}\underline{93.25\%} & \cellcolor{orange!30}\textbf{96.31\%} & 67.58\% \\
 \cline{3-7}
 &  & Overall & \cellcolor{yellow!30}\underline{63.22\%} & -1703.38\% & -195.34\% & \cellcolor{orange!30}\textbf{67.06\%} \\
 \bottomrule
\end{NiceTabular}
\end{table}

\begin{figure}[htbp]
    \centering
    \input{supplement/diagrams/grids/synth/diagram}
    \caption{Qualitative comparison of artifact removal (top row) and artifact detection (bottom row) on different kinds of synthetic artifacts.}
    \label{fig:ab-synth-artifact-detection-removal}
\end{figure}

\begin{figure}[htbp]
    \centering
    \input{supplement/diagrams/grids/real/diagram}
    \caption{Comparison of artifact removal on real-world recordings. Zoom-ins (red border) provided for better visual assessment.}
    \label{fig:ab-real-artifact-removal}
\end{figure}

The ablation results highlight three major factors influencing the performance and stability of \textsc{DeLux}: (i) the contribution of event data, (ii) the role of the detector, and (iii) the effect of the non-mask loss.

\paragraph{Impact of Event Data.}
Removing the event channel (\textsc{DeLux-NE}) leads to a clear drop in both detection and reconstruction quality. In synthetic artifacts, models without events produce coarse, incomplete artifact maps and fail to recover the image structure when glare, flare, or overexposure occlude large regions. In real recordings, they also exhibit fewer catastrophic failures, but only because the model lacks the event-driven cues that normally push it to correct severe degradations. In general, event information is essential for accurate localization and recovery of structure in saturated regions.

\paragraph{Role of the Detector.}
Disabling the detector (\textsc{DeLux-ND}) forces the reconstruction network to operate without spatial guidance. This configuration consistently performs worst among all tested variants on synthetic data and loses interpretability due to the absence of explicit artifact maps. The results confirm that providing a dedicated artifact-location estimate is critical for targeted removal.

\begin{figure}[htbp]
    \centering
    \begin{subfigure}[t]{0.48\linewidth}
        \centering
        \includegraphics[width=0.9\linewidth]{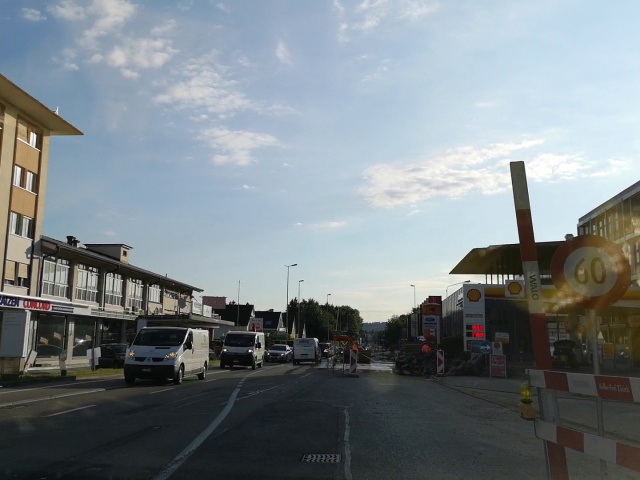}
        \captionsetup{width=0.9\linewidth}
        \caption{RGB Input Frame}
    \end{subfigure}
    \begin{subfigure}[t]{0.48\linewidth}
        \centering
        \includegraphics[width=0.9\linewidth]{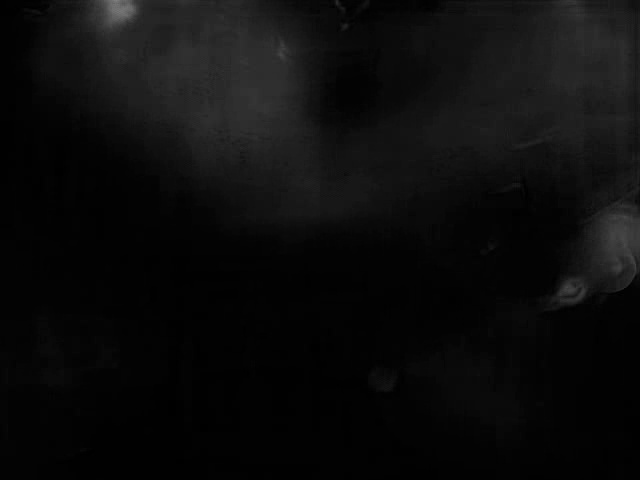}
        \captionsetup{width=0.9\linewidth}
        \caption{Detector Output on Input Frame}
    \vspace{0.25cm}
    \end{subfigure}
     \begin{subfigure}[t]{0.48\linewidth}
        \centering
        \includegraphics[width=0.9\linewidth]{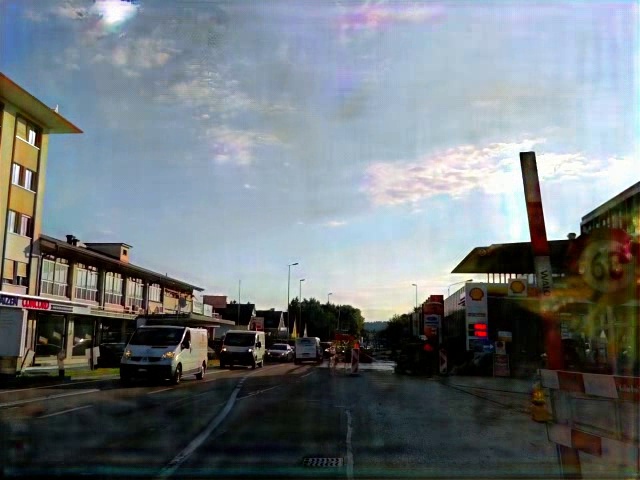}
        \captionsetup{width=0.9\linewidth}
        \caption{RGB Artifact Remover Output}
    \end{subfigure}
    \begin{subfigure}[t]{0.48\linewidth}
        \centering
        \includegraphics[width=0.9\linewidth]{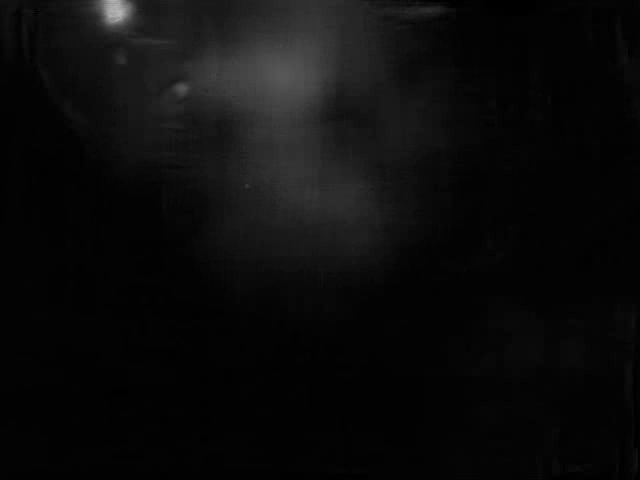}
        \captionsetup{width=0.9\linewidth}
        \caption{Detector Output on the Output of the Remover}
    \end{subfigure}
    \caption{Failure case for the \textsc{DeLux-D} model on a frame from the \textit{sun6} recording (a). The detector confuses clouds as a light artifact (b), which causes a false indication to the remover to make an adjustment to the input image. The resulting image is degraded (c) and has a new artifact in place of the initial detection. The secondary detection (d) on the resulting image shows more artifacts than the input.}
    \label{fig:disc_failure}
\end{figure}

\begin{figure}[htbp]
    \centering
    \begin{subfigure}[t]{0.48\linewidth}
        \centering
        \includegraphics[width=0.9\linewidth]{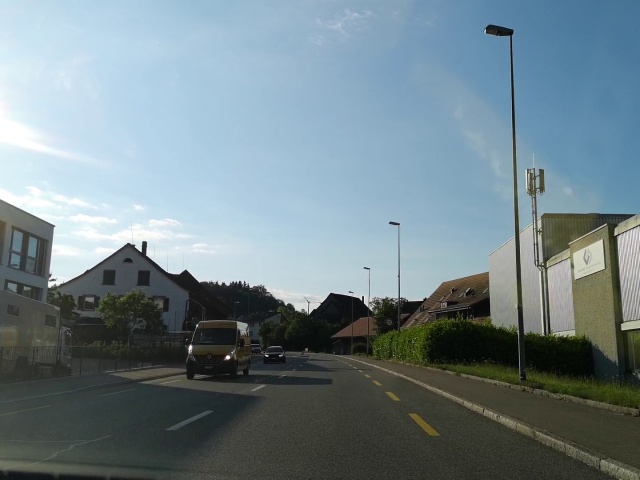}
    \end{subfigure}
    \begin{subfigure}[t]{0.48\linewidth}
        \centering
        \includegraphics[width=0.9\linewidth]{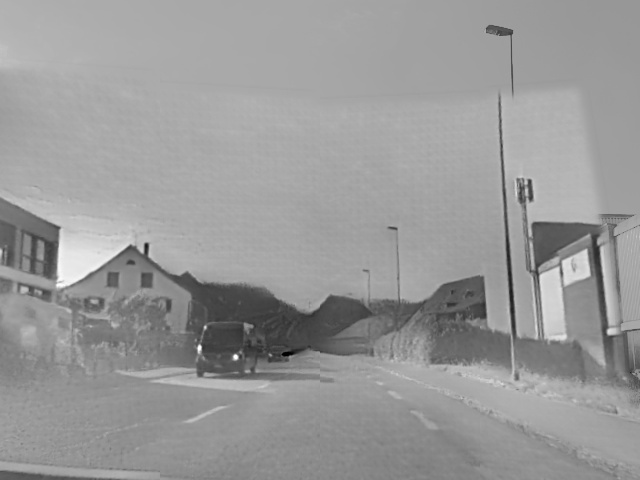}
    \end{subfigure}
    \caption{Sample frame from the \textit{sun10} recording, which was excluded from evaluation due to poor alignment quality. The RGB frame (left) is visibly misaligned with the reconstructed event frame (right) after infilling. The misalignment is particularly evident around the lamp post, highlighting limitations in our alignment process.
}
    \label{fig:disc_misalignment}
\end{figure}

\paragraph{Failure Modes and Data Quality.}
The most severe failure cases arise from noisy or poorly aligned event reconstructions. Misalignment between RGB and event frames can mislead the detector and propagate to the remover, occasionally introducing new artifacts. An example of such failure is presented in Figure~\ref{fig:disc_failure}. These failures are closely related to recordings aligned using our calibration-free procedure, suggesting that improvements in event-to-RGB alignment (an example of bad alignment is shown in Figure~\ref{fig:disc_misalignment}) or larger volumes of perfectly aligned synthetic data—would further strengthen the system.

\paragraph{Effect of the Non-mask Loss.}
Disabling $\mathcal{L}_{\text{nm}}$ (\textsc{DeLux-D}) increases detection accuracy but leads to overly aggressive, poorly localized edits during removal. In contrast, enabling the loss stabilizes the reconstructor, constraining modifications to predicted artifact regions, and producing outputs that better align with manually annotated masks. This makes non-mask loss preferable when balancing detection quality with controlled reconstruction.

\begin{figure}[htbp]
    \centering
    \begin{subfigure}[t]{0.24\linewidth}
        \centering
        \includegraphics[width=0.95\linewidth]{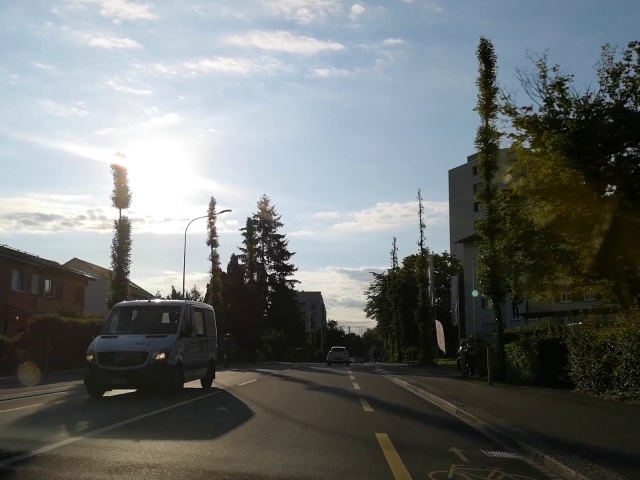}
        \captionsetup{width=0.9\linewidth}
        \caption{RGB Input Frame}
    \end{subfigure}
    \begin{subfigure}[t]{0.24\linewidth}
        \centering
        \includegraphics[width=0.95\linewidth]{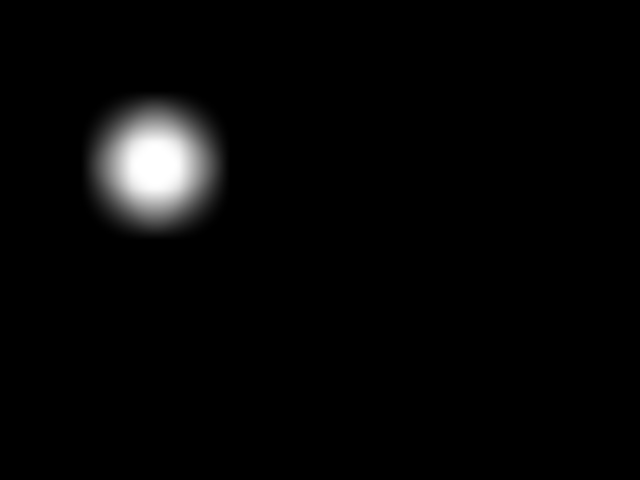}
        \captionsetup{width=0.9\linewidth}
        \caption{Manually Annotated Soft Mask}
    \end{subfigure}
     \begin{subfigure}[t]{0.24\linewidth}
        \centering
        \includegraphics[width=0.95\linewidth]{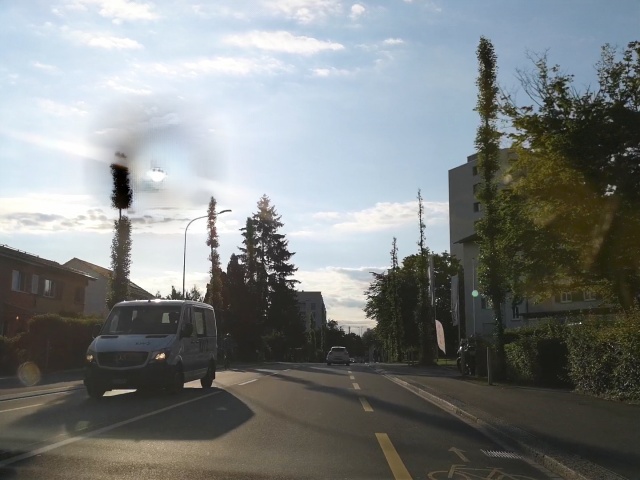}
        \captionsetup{width=0.9\linewidth}
        \caption{DeLux-D}
    \end{subfigure}
    \begin{subfigure}[t]{0.24\linewidth}
        \centering
        \includegraphics[width=0.95\linewidth]{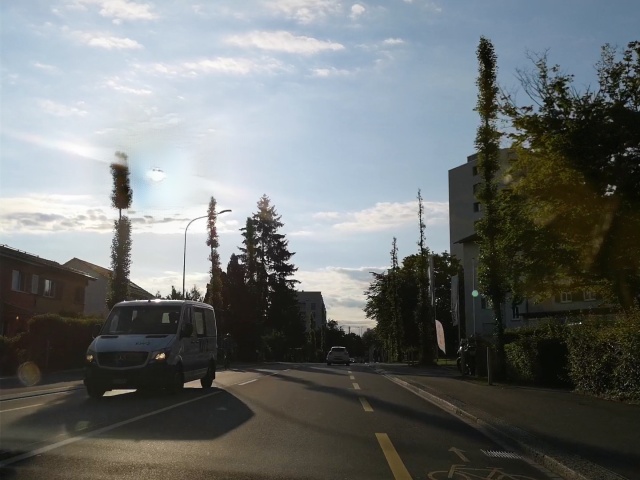}
        \captionsetup{width=0.9\linewidth}
        \caption{DeLux}
    \end{subfigure}
    \caption{Comparison of the model outputs on a sample frame from \textit{sun2} recording (a), when provided with a blurred manually annotated mask (b), between non-mask loss $\mathcal{L}_{\text{nm}}$ disabled (c) and enabled (d). While both models successfully remove the artifact within the masked area, the \textsc{Delux-D} variant (c) introduces noticeable discoloration and exhibits greater sensitivity to the intensity of the input mask.}
    \label{fig:disc_map_study_manual}
\end{figure}

\begin{figure}[htbp]
    \centering
    \begin{subfigure}[t]{0.24\linewidth}
        \centering
        \includegraphics[width=0.95\linewidth]{supplement/figures/mask_study_input.jpg}
        \captionsetup{width=0.9\linewidth}
        \caption{RGB Input Frame}
    \end{subfigure}
    \begin{subfigure}[t]{0.24\linewidth}
        \centering
        \includegraphics[width=0.95\linewidth]{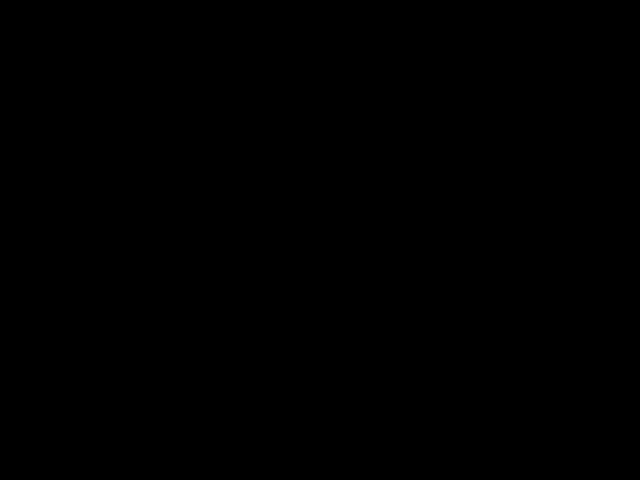}
        \captionsetup{width=0.9\linewidth}
        \caption{Empty Input Artifact Mask}
    \end{subfigure}
     \begin{subfigure}[t]{0.24\linewidth}
        \centering
        \includegraphics[width=0.95\linewidth]{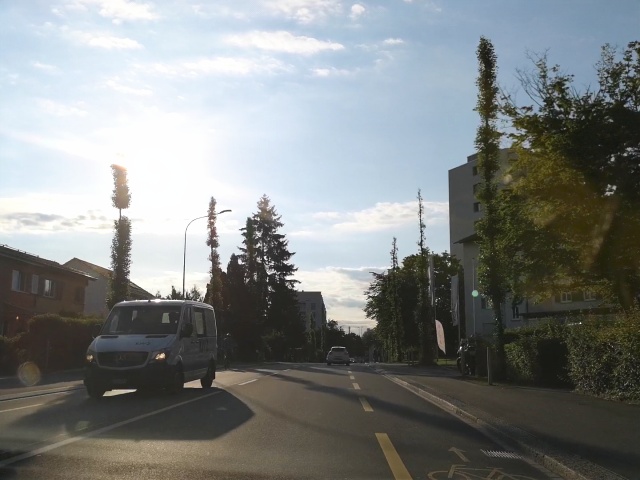}
        \captionsetup{width=0.9\linewidth}
        \caption{DeLux-D}
    \end{subfigure}
    \begin{subfigure}[t]{0.24\linewidth}
        \centering
        \includegraphics[width=0.95\linewidth]{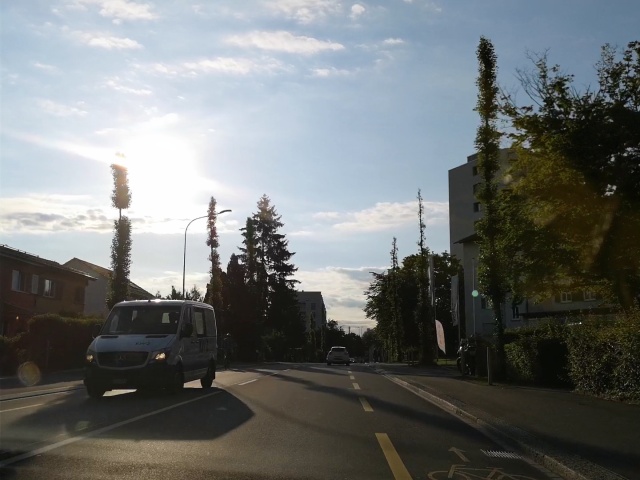}
        \captionsetup{width=0.9\linewidth}
        \caption{DeLux}
    \end{subfigure}
    \caption{Comparison of the model outputs on a sample frame from \textit{sun2} recording (a), when provided with an empty mask (b), between non-mask loss $\mathcal{L}_{\text{nm}}$ disabled (c) and enabled (d). Both models do not remove the artifacts, but the baseline model slightly increases the overall brightness of the image (c), while the model trained with the non-mask loss term leaves the image unchanged (d).}
    \label{fig:disc_map_study_none}
\end{figure}

The $\mathcal{L}_{\text{nm}}$ loss also improves the consistency of the model across different sources of artifact maps. As shown in Figure~\ref{fig:disc_map_study_manual}, the baseline \textsc{DeLux} handles manually provided soft masks more reliably, whereas \textsc{DeLux-D} introduces color shifts and over-edits even when the annotated region is accurate. A similar pattern appears when an empty mask is given (Figure~\ref{fig:disc_map_study_none}): \textsc{DeLux} preserves the input image, while \textsc{DeLux-D} still applies unwanted modifications despite the absence of any indicated artifact regions.

\medskip
In summary, the ablations demonstrate that (1) events and the detector are both essential components of \textsc{DeLux}, (2) moderate rule-based fusion can enhance performance in challenging overexposed regions, and (3) the non-mask loss stabilizes removal by constraining the model to operate within predicted artifact regions.

\subsection{Robustness of the SAS Metric}
\label{sec:supplement-sas-robustness}

Because Strong Artifact Suppression (SAS) is computed from a learned artifact detector, its absolute magnitude depends on the detector used. It is therefore best read as a \emph{relative} artifact-suppression indicator for ranking methods rather than as an absolute score. To verify that the ranking is not an artifact of using our own \textsc{DeLux-D} detector, we recompute $\Delta$\,SAS using detections from the independently trained DAD\textsuperscript{\textdagger}
model (Table~\ref{tab:dad-sas}). Under this independent detector, \textsc{DeLux} still attains the best overall score and wins on 5 of the 8 sequences, confirming that the ordering is stable across detectors.

\begin{table}[h]
\centering
\caption{$\Delta$ SAS $\uparrow$ based on detection outputs from DAD$^\dagger$ model.}
\label{tab:dad-sas}
\resizebox{\linewidth}{!}{
\begin{tabular}{llcccccc}
\toprule
   & DAD\textsuperscript{\textdagger} & F7K & Wu et al.  & SHDR & HDRev-Diff & DeLux \\
\midrule
 Sun 02 & \cellcolor{yellow!30}{79.35\%} & 34.07\% & -428.94\% & -453.37\% & -5260.29\% & \cellcolor{orange!30}{79.42\%} \\
  Sun 05 & -184.08\% & \cellcolor{yellow!30}{17.12\%} & -268.66\% & -260.29\% & -271.06\% & \cellcolor{orange!30}{81.86\%} \\
  Sun 06 & 27.51\% & 8.99\% & \cellcolor{yellow!30}{59.13\%} & -125.79\% & -109341.52\% & \cellcolor{orange!30}{64.49\%} \\
  Sun 09 & \cellcolor{yellow!30}{64.99\%} & -149.32\% & 27.84\% & -373.00\% & -3477.06\% & \cellcolor{orange!30}{73.72\%} \\
 Sun 10 & \cellcolor{yellow!30}{72.37\%} & 27.56\% & -72.01\% & -267.06\% & -15016.77\% & \cellcolor{orange!30}{79.45\%} \\
  Sun 11 & -278.51\% & \cellcolor{orange!30}{4.97\%} & -6319.55\% & -324.93\% & -816645.71\% & \cellcolor{yellow!30}{-116.01\%} \\
  Zurich City 01-e & \cellcolor{orange!30}{29.75\%} & \cellcolor{yellow!30}{-43.39\%} & -266.31\% & -1037.68\% & -1542.28\% & -90.59\% \\
  Zurich City 12-a & \cellcolor{orange!30}{59.14\%} & \cellcolor{yellow!30}{28.22\%} & -210.44\% & -3676.72\% & -2694.57\% & 24.55\% \\
\cline{1-7}
  Overall & -14.51\% & \cellcolor{yellow!30}{-3.96\%} & -1027.44\% & -663.89\% & -133185.49\% & \cellcolor{orange!30}{15.54\%} \\
\bottomrule
\end{tabular}}
\end{table}

\section{Additional Artifact Detection Results}
\label{sec:supplement-detection}

\begin{figure}[htb]
    \centering
    \input{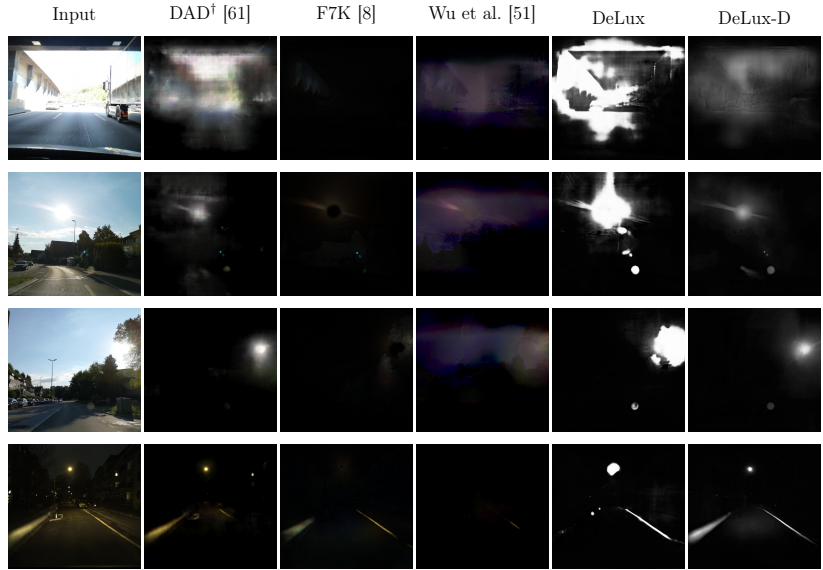}
    \caption{Comparison of artifact detection on selected frames from real-world recordings.}
    \label{fig:ab-real-artifact-detection}
\end{figure}

This section includes additional artifact detection results for \textsc{DeLux} and compared baselines (Figure~\ref{fig:ab-real-artifact-detection}).

\section{Additional Artifact Removal Results}
\label{sec:supplement-removal}

\begin{figure*}[htbp]
    \centering
    \input{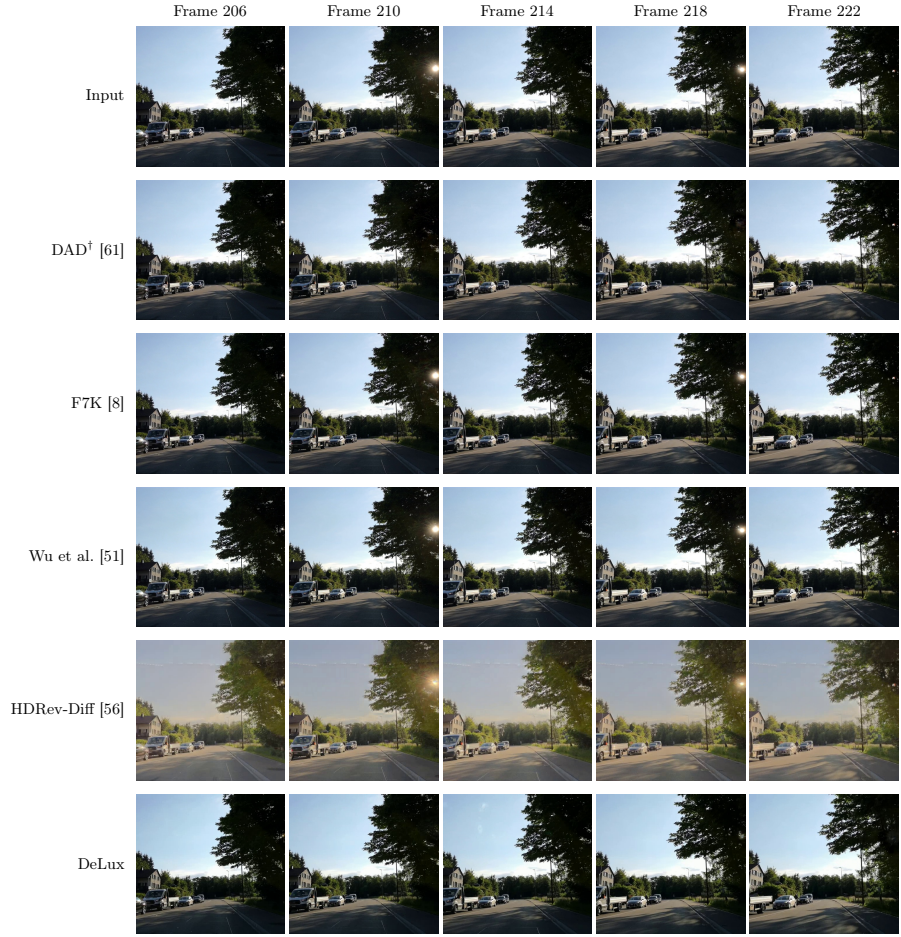}
    \caption{
        Example of flicker artifact removal on five frames from the \textit{sun11} video recording, spaced by $150ms$. Visible glare on frames $210$ and $218$, lack of it in the remaining frames, which implies a flicker artifact.
    }
    \label{fig:supp-flicker}
\end{figure*}

This section includes additional artifact removal results for \textsc{DeLux} and compared baselines (Figure~\ref{fig:supp-flicker}). 

In addition to the flicker removal examples, we provide an exemplar recording (\textit{sun2.mp4}) in the supplementary materials. This recording includes the processed video, its corresponding detection map, and the reconstructed output generated by our model. The example illustrates how \textsc{DeLux} identifies artifact-prone regions and reconstructs them in a temporally consistent manner. Reviewers are referred to the supplementary video for a qualitative demonstration of the model's behavior beyond the static frames shown in this section.

\section{Fusion Module Details}
\label{sec:supplement-fusion}

\begin{figure*}[ht]
\centering
\begin{subfigure}[t]{0.9\textwidth}
	\centering
	\begin{minipage}[t]{0.2\textwidth}
		\vspace{1cm}
		\resizebox{\textwidth}{!}{%
		\begin{minipage}{2.5cm}
		\raggedright
		{\scriptsize \textbf{Legend:}}\\[0.1em]
		\renewcommand{\arraystretch}{0.5}
        \tiny
		\begin{tabular}{@{}l@{}}
			\texttt{\colorbox{gray!30}{CB~}} Conv Block \\
			\texttt{\colorbox{blue!20}{UCB}} UpConv Block \\
			\texttt{\colorbox{red!30}{MP~}} MaxPool 2D \\
			\texttt{\colorbox{green!30}{BN~}} 2D Batch Normaliztion \\
			\texttt{\colorbox{darkgray!30}{C~~}} ($k{\times}k$) 2D Convolution \\
			\texttt{\colorbox{cyan!30}{TC~}} ($k{\times}k$) Trans. 2D Conv. \\
			\texttt{\colorbox{orange!30}{BI~}} Bilinear Interpolation \\
		\end{tabular}
		\end{minipage}%
		}%
	\end{minipage}
	\begin{minipage}[t]{0.73\textwidth}
		\centering
		{\scriptsize \textbf{U-Net Architecture}}
		\vspace{0.3em}
		
		\resizebox{\textwidth}{!}{%
\footnotesize
\begin{tikzpicture}[ node distance=1.5cm ]
\tikzstyle{layer} = [rectangle, text width = 1.5cm, text centered, draw = black, fill = gray! 30, minimum width=1cm, minimum height = 0.5cm]
\tikzstyle{upconv} = [fill=blue! 20]
\tikzstyle{activation} = [fill = yellow! 30]
\tikzstyle{maxpool} = [fill = red! 30]
\tikzstyle{skip} = [draw=gray, thick, dashed, ->]
\tikzstyle{flow} = [->, thick]

\node (det_l1) [layer, rotate=90, minimum width=3cm] { CB };
\node (input) [left of=det_l1, text width = 1cm] {Input\\Image};
\node (det_a1) [layer, right of=det_l1, maxpool, rotate=90, yshift=1cm, minimum width=3cm] { MP };

\node (det_dot_1) [right of=det_l1] {$\dots$};
\node (det_l2) [layer, right of=det_dot_1, rotate=90, minimum width=2cm, yshift=0.5cm, xshift=-0.5cm] { CB };
\node (det_a2) [layer, right of=det_l2, maxpool, rotate=90, yshift=1.5cm, minimum width=2cm, yshift=-0.5cm] { MP };

\node (det_l3) [layer, right of=det_l2, rotate=90, minimum width=1cm, text width=1cm, xshift=-0.35cm] { CB };
\node (det_a3) [layer, right of=det_l3, maxpool, rotate=90, yshift=1cm, minimum width=1cm, text width=1cm] { MP };

\node (det_l4) [layer, upconv, right of=det_l3, rotate=90, minimum width=2cm, xshift=0.3cm] { UCB };

\node (det_dot_2) [right of=det_l4, yshift=0.5cm] {$\dots$};

\node (det_l5) [layer, upconv, right of=det_dot_2, rotate=90, minimum width=3cm, yshift=0.5cm] { UCB };

\node (det_l6) [layer,  right of=det_l5, rotate=90, minimum width=3cm, yshift=0.5cm, text width=2cm] { C($1\times1$)};
\node (det_a6) [layer, right of=det_l6, activation, rotate=90, yshift=1cm, minimum width=3cm] { Sigmoid };

\node (output) [right of=det_a6, text width=1cm] {Output\\Image};

\draw[decorate, decoration={brace, mirror, amplitude=10pt}] (-0.5,-1.55) -- (3.5,-1.55) node[midway, yshift=-20pt] {$n$ blocks};

\draw[decorate, decoration={brace, mirror, amplitude=10pt}] (5,-1.55) -- (8.5,-1.55) node[midway, yshift=-20pt] {$n$ blocks};

\draw[flow] (input) -- (det_l1);
\draw[flow] (det_a1) -- (det_dot_1);
\draw[flow] (det_dot_1) -|- (det_l2);
\draw[flow] (det_a2) -|- (det_l3);
\draw[flow] (det_a3) -|- (det_l4);
\draw[flow] (det_l4) -|- (det_dot_2);
\draw[flow] (det_dot_2) -- (det_l5);
\draw[flow] (det_l5) -- (det_l6);
\draw[flow] (det_a6) -- (output);

\draw[skip] (0,1.5) -- (0,2) -- (8,2)  node[midway, yshift=-0.7cm] {Skip Connections} -- (8,1.5);
\draw[skip] (2.5,0.5) -- (2.5,0.75) -- (5.5,0.75) -- (5.5,0.5);

\end{tikzpicture}
}%
		
		\vspace{1em}
		
		\begin{minipage}[t]{0.48\textwidth}
			\centering
			{\scriptsize \textbf{Conv Block}}
			\vspace{0.3em}
			
			\resizebox{\linewidth}{!}{%
\begin{tikzpicture}[ node distance=1.5cm ]

\Large

\tikzstyle{add} = [circle, draw=black, minimum size=0.7cm, inner sep=0pt]
\tikzstyle{skip} = [draw=gray, thick, dashed, ->]
\tikzstyle{flow} = [->, thick]
\tikzstyle{layer} = [rectangle, text width = 2cm, text centered, draw = black, fill = darkgray! 30, minimum width=2cm, minimum height = 0.5cm, rotate=90]
\tikzstyle{activation} = [fill = yellow! 30]
\tikzstyle{bn} = [fill = green! 30]
\tikzstyle{add} = [circle, draw=black, minimum size=0.7cm, inner sep=0pt]

\node (input) [text width = 1cm] {Input};

\node (conv1) [layer, below of=input] { C($k\times k$) };
\node (act1) [layer, activation, below of=input, yshift=-0.7cm] {ReLU};
\node (conv2) [layer, below of=conv1, yshift=-0.5cm] { C($k\times k$) };
\node (act2) [layer, activation, below of=conv1, yshift=-1.2cm] { ReLU };
\node (dots) [right of=conv2, xshift=0.45cm] {$\dots$};
\node (conv3) [layer, below of=dots] { C($k\times k$) };
\node (act3) [layer, activation, below of=dots, yshift=-0.7cm] { ReLU };
\node (conv_skip) [layer, left of=conv1, xshift=-1.25cm] { C($1\times 1$) };
\node (add) [add, right of=conv_skip, xshift=3cm] {$+$};
\node (bnorm) [layer, bn, below of=add] {BN};
\node (act_end) [layer, activation, below of=add, yshift=-0.7cm] {ReLU};
\node (output) [right of=act_end, text width=1cm] {Output};

\draw[decorate, decoration={brace, amplitude=10pt}] ([xshift=-15pt]conv2.east) -- ([xshift=15pt]act3.east) node[midway, yshift=18pt] {$m$ layers};

\draw [flow] (input) -- (conv1);
\draw [flow] (act1) -- (conv2);
\draw [flow] (act2) -- (dots);
\draw [flow] (dots) -- (conv3);
\draw [flow] (act3.south) -| ([yshift=-0.4cm, xshift=0.4cm]act3.south west) -| (add);
\draw [flow] (add) -- (bnorm);
\draw [flow] (act_end) -- (output);

\draw [flow] (input) |- (conv_skip);
\draw [flow] (conv_skip) -- (add);

\end{tikzpicture}

}%
		\end{minipage}\hfill
		\begin{minipage}[t]{0.48\textwidth}
			\centering
			{\scriptsize \textbf{UpConv Block}}
			\vspace{0.3em}
			
			\resizebox{\linewidth}{!}{%
\begin{tikzpicture}[ node distance=1.5cm ]
\large

\tikzstyle{add} = [circle, draw=black, minimum size=0.7cm, inner sep=0pt]
\tikzstyle{skip} = [draw=gray, thick, dashed, ->]
\tikzstyle{flow} = [->, thick]
\tikzstyle{layer} = [rectangle, text width = 2cm, text centered, draw = black, fill = gray! 30, minimum width=2cm, minimum height = 0.5cm, rotate=90]
\tikzstyle{convt} = [fill = cyan! 20]
\tikzstyle{activation} = [fill = yellow! 30]
\tikzstyle{interpolation} = [fill = orange! 30]
\tikzstyle{concat} = [circle, draw=black, minimum size=0.7cm, inner sep=0pt]

\node (skipinput) [text width = 1cm] { Skip Input };
\node (upinput) [text width = 1cm, below of=skipinput, yshift=-1.25cm] {Up Input};

  \node (tconv) [layer, convt, below of=upinput] {TC($k\times k$)};
\node (tact) [layer, activation, below of=upinput, yshift=-0.5cm] {ReLU};

\node (itp) [layer, interpolation, below of=tact] {BI};
\node (concat) [concat, label=left:Concat., right of=itp, yshift=1.5cm] {\Large $\times$};

\node (convblock) [layer, below of=concat] {CB};

\node (output) [text width = 1cm, right of=convblock] { Output };

\draw [skip] (skipinput) -| (concat);
\draw [flow] (upinput) -- (tconv);
\draw [flow] (tact) -- (itp);
\draw [flow] (itp) -| (concat);
\draw [flow] (concat) -- (convblock);
\draw [flow] (convblock) -- (output);

\end{tikzpicture}
}%
		\end{minipage}
	\end{minipage}
\end{subfigure}

\vspace{0.8em}
\noindent\rule{\textwidth}{0.4pt}
\vspace{0.8em}

\begin{subfigure}[t]{0.9\textwidth}
	\centering
	{\scriptsize \textbf{Training Pipeline}}
	\vspace{0.3em}
	
	\input{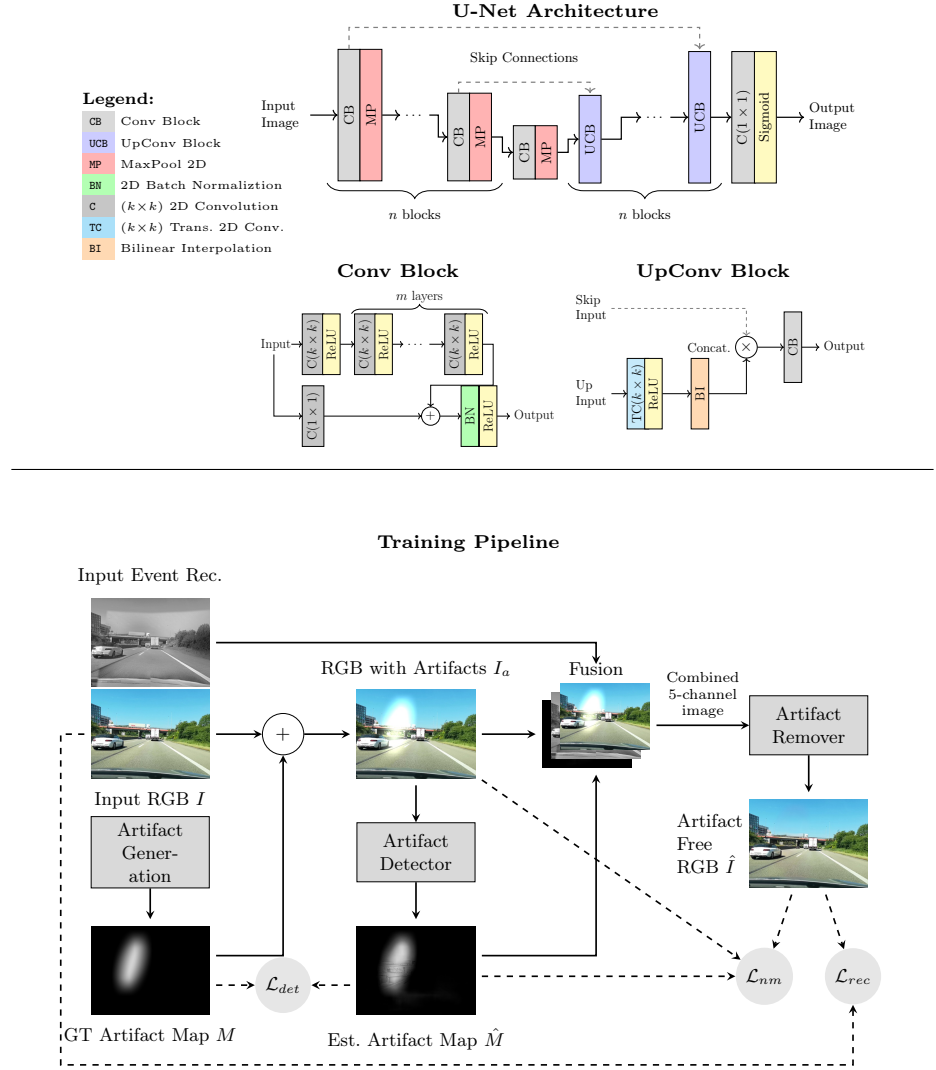}
\end{subfigure}

\caption{A detailed depiction of the \textsc{DeLux} architecture, highlighting the multimodal fusion stage that blends the RGB, event-reconstruction, and artifact-mask channels.}
\label{fig:supp-fusion}
\end{figure*}

This section expands the description of the multimodal fusion stage. The module takes the three spatially aligned representations (the RGB frame, the event-based grayscale reconstruction, and the soft artifact mask) and concatenates them along the channel dimension before passing them through a stack of convolutional layers (Fig.~\ref{fig:supp-fusion}). It does not use any attention, but instead, the soft artifact mask is exposed as a manual-override interface. This makes fusion behavior configurable at inference \emph{without retraining}: supplying an edited or empty mask predictably changes which regions are modified, as shown in the mask-conditioning studies (Figs.~\ref{fig:disc_map_study_manual} and~\ref{fig:disc_map_study_none}).

\section{Computational Cost}
\label{sec:supplement-cost}

Table~\ref{tab:benchmark} reports model size and inference speed, measured on a 100-frame sample with a 50-frame warmup on an NVIDIA A100 GPU and an AMD EPYC CPU. \textsc{DeLux} runs at 16.5\,FPS with 90.2\,M parameters---an order of magnitude smaller and faster than the diffusion-based HDRev-Diff baseline (1.6\,B parameters, 1.1\,FPS), while remaining competitive with the lightweight single-network baselines.

\begin{table}[htbp]
\centering
\caption{Model sizes and inference speed.}
\label{tab:benchmark}
\begin{tabular}{lrrrrrr}
\toprule
\multirow{2}{*}{\textbf{Model}} & \multirow{2}{*}{\textbf{Parameters (M)}} & \multirow{2}{*}{\textbf{Size (MB)}}
  & \multicolumn{2}{c}{\textbf{GPU}} & \multicolumn{2}{c}{\textbf{CPU}} \\
\cmidrule(lr){4-5}\cmidrule(lr){6-7}
 &  &  & \textbf{Latency (ms)} & \textbf{FPS} & \textbf{Latency (s)} & \textbf{FPS} \\
\midrule
DeLux       & 90.2      & 344.2     & 60.6       & 16.5     & 13.7      & 0.07 \\
DAD$^\dagger$         & 41.8      & 159.6     & 2.3        & 431.1    & 0.2       & 4.88 \\
F7K         & 20.5      & 78.7      & 64.2       & 15.6     & 2.3       & 0.43 \\
HDRev-Diff  & 1615.2    & 6161.5    & 944.8      &  1.1     & 8.8       & 0.11  \\
SHDR        & 29.0      & 344.0     & ---        & ---      & 6.0       & 0.16 \\
Wu et al.   & 31.0      & 118.9     & ---        &  ---     & 2.3       & 0.43  \\
\bottomrule
\end{tabular}
\end{table}

\end{document}